%% file: submission_7969679-20260905-0413/sample-sigplan.tex
\documentclass[sigconf]{acmart}

\usepackage{booktabs,diagbox}
\usepackage{enumitem}
\usepackage{multirow}
\usepackage{caption}
\usepackage{array}
\usepackage{makecell}
\usepackage{color}
\usepackage{xcolor}

\usepackage{subcaption}
\usepackage{algorithm}

\usepackage{pifont}

\usepackage{threeparttable}
\usepackage{pdfpages}
\usepackage{pdfcomment}
\usepackage{tikz}
\usepackage{amsmath}
\usepackage{float}
\usepackage{bm}
\usepackage{listings} % 代码展示
\usepackage{xcolor} % 使用颜色
\usepackage{stfloats}
\usepackage{adjustbox}
\usepackage{graphicx,booktabs}
\usepackage{booktabs,threeparttable}

\usepackage{xcolor}
\usepackage{booktabs}
\usepackage{tabularx}
\usepackage{fontspec}
\definecolor{matchingcolor}{HTML}{0072B2}
\definecolor{qacolor}{HTML}{D55E00}

\acmConference[KDD '27]
{Proceedings of the 33rd ACM SIGKDD Conference on Knowledge Discovery and Data Mining}
{August 1--5, 2027}
{San Jose, CA, USA}

\acmYear{2027}
\copyrightyear{2027}
\renewcommand\footnotetextcopyrightpermission[1]{} % removes footnote with conference information in first column 

\newcommand{\shortname}{\textsc{CLLPU}} \newcommand{\fullname}{\textbf{\underline{C}}ross-\textbf{\underline{L}}ingual and \textbf{\underline{L}}anguage-Bound \textbf{\underline{P}}rotocol for LLM \textbf{\underline{U}}nlearning}

\begin{document}

%%
%% The "title" command has an optional parameter,
%% allowing the author to define a "short title" to be used in page headers.
\title{Beyond Cross-Lingual Transfer: Benchmarking Propagation Boundaries in Multilingual LLM Unlearning}

%%
%% The "author" command and its associated commands are used to define
%% the authors and their affiliations.
%% Of note is the shared affiliation of the first two authors, and the
%% "authornote" and "authornotemark" commands
%% used to denote shared contribution to the research.

\author{Pengyang Shao}
% \authornote{Equal Contribution.}
\affiliation{%
 \institution{National University of Singapore}
 \country{Singapore}
}
% \email{shaopymark@gmail.com}

\author{Chuanpeng Lu}
% \authornotemark[1]
\affiliation{%
 \institution{Hefei University of Technology}
 \state{Anhui}
 \country{China}
}

\author{Wei Qin}
% \authornote{Corresponding author.}
\affiliation{%
 \institution{Hefei University of Technology}
 \state{Anhui}
 \country{China}
}

\author{Yanzheng Jin}
% \authornotemark[1]
\affiliation{%
 \institution{National University of Singapore}
 \country{Singapore}
}

\author{Xiaohao Liu}
\affiliation{%
 \institution{National University of Singapore}
 \country{Singapore}
}

\author{Xi Ai}
\affiliation{%
 \institution{National University of Singapore}
 \country{Singapore}
}

\author{Kenji Kawaguchi}
\affiliation{%
 \institution{National University of Singapore}
 \country{Singapore}
}

\author{Richang Hong}
\affiliation{%
 \institution{Hefei University of Technology}
 \state{Anhui}
 \country{China}
}

%%
%% By default, the full list of authors will be used in the page
%% headers. Often, this list is too long, and will overlap
%% other information printed in the page headers. This command allows
%% the author to define a more concise list
%% of authors' names for this purpose.
\renewcommand{\shortauthors}{Shao et al.}

%%
%% The abstract is a short summary of the work to be presented in the
%% article.
\input{tex/0abstract}   

%%
%% The code below is generated by the tool at http://dl.acm.org/ccs.cfm.
%% Please copy and paste the code instead of the example below.
%%
% \begin{CCSXML}
% <ccs2012>
%  <concept>
%   <concept_id>00000000.0000000.0000000</concept_id>
%   <concept_desc>Do Not Use This Code, Generate the Correct Terms for Your Paper</concept_desc>
%   <concept_significance>500</concept_significance>
%  </concept>
%  <concept>
%   <concept_id>00000000.00000000.00000000</concept_id>
%   <concept_desc>Do Not Use This Code, Generate the Correct Terms for Your Paper</concept_desc>
%   <concept_significance>300</concept_significance>
%  </concept>
%  <concept>
%   <concept_id>00000000.00000000.00000000</concept_id>
%   <concept_desc>Do Not Use This Code, Generate the Correct Terms for Your Paper</concept_desc>
%   <concept_significance>100</concept_significance>
%  </concept>
%  <concept>
%   <concept_id>00000000.00000000.00000000</concept_id>
%   <concept_desc>Do Not Use This Code, Generate the Correct Terms for Your Paper</concept_desc>
%   <concept_significance>100</concept_significance>
%  </concept>
% </ccs2012>
% \end{CCSXML}

% \ccsdesc[500]{Do Not Use This Code~Generate the Correct Terms for Your Paper}
% \ccsdesc[300]{Do Not Use This Code~Generate the Correct Terms for Your Paper}
% \ccsdesc{Do Not Use This Code~Generate the Correct Terms for Your Paper}
% \ccsdesc[100]{Do Not Use This Code~Generate the Correct Terms for Your Paper}

%%
%% Keywords. The author(s) should pick words that accurately describe
%% the work being presented. Separate the keywords with commas.
%% A "teaser" image appears between the author and affiliation
%% information and the body of the document, and typically spans the
%% page.

% \received{20 February 2007}
% \received[revised]{12 March 2009}
% \received[accepted]{5 June 2009}

%%
%% This command processes the author and affiliation and title
%% information and builds the first part of the formatted document.
\maketitle

\input{tex/1introduction}

\input{tex/2related_work}

\input{tex/3construction_simplified}

\input{tex/3method}

\input{tex/4experiment}

\input{tex/5conclusion}

\newpage

\bibliographystyle{ACM-Reference-Format}
\balance
\bibliography{references}

% \newpage
% \let\addcontentsline\OriginalAddContentsLine
% \setcounter{tocdepth}{2}
\appendix
% \tableofcontents
\input{tex/6appendix}

\end{document}

%% file: tex/0abstract.tex
\begin{abstract}
Large Language Model (LLM) unlearning aims to suppress target knowledge while preserving general capabilities. In multilingual settings, unlearning must additionally propagate within its intended linguistic scope. However, existing evaluations mainly measure cross-lingual transfer and cannot distinguish insufficient from excessive propagation. 
We introduce \textbf{CLLPU} (Cross-Lingual and Language-Bound Protocol for LLM Unlearning), a multilingual benchmark that formulates this problem through two settings: common-goal forgetting, where target knowledge should be suppressed across all languages, and language-conditioned forgetting, where suppression should remain confined to a designated language. CLLPU combines goal-guided topic pairing, schema-aware relation matching, and dual-anchor multilingual translation to construct 800 matched knowledge-unit pairs and 72,000 QA instances across ten languages. Experiments with six representative methods on Llama-3.1-8B-Instruct reveal opposite failure modes: forgetting remains incomplete when universal suppression is required, yet spreads beyond the intended boundary when language-conditioned confinement is required. We further find that general multilingual utility can conceal damage to neighbor knowledge. These findings establish propagation control as a central challenge for multilingual LLM unlearning. We publicly release CLLPU together with its construction pipeline\footnote{https://github.com/CLLPU/CLLPU-bench}.
\end{abstract}

%% file: tex/1introduction.tex
\section{Introduction}
Large Language Model (LLM) unlearning seeks to remove the influence of private, copyrighted, or otherwise sensitive data from a deployed LLM while preserving its general capabilities~\cite{zhang2025rule,zhai2026maximizing,gao2025large,yangexploring,liu2025rethinking}.
This need is further reinforced by practical and regulatory demands, such as the right to erasure under the General Data Protection Regulation (GDPR)~\cite{protection2018general} and deletion rights under the California Consumer Privacy Act (CCPA)~\cite{pardau2018california}.

Among LLM unlearning research topics, how to reliably evaluate an unlearned LLM remains a fundamental problem~\cite{dorna2026openunlearning,yoon2025r,shi2025muse}. Existing benchmarks operationalize this objective by separately evaluating the removal of designated target knowledge and the preservation of non-target capabilities, e.g., TOFU examines whether a model forgets selected fictitious author profiles while retaining those of the remaining authors~\cite{maini2024tofu}. Under this paradigm, successful unlearning is characterized by a substantial reduction in target-knowledge accessibility, together with minimal degradation in non-target performance. 

Multilingual settings introduce a further requirement for unlearning evaluation: forgetting should propagate within its intended linguistic scope~\cite{lu2025learn,choi2024cross,lizzo2026evaluating}. Recent studies investigate cross-lingual forgetting by applying unlearning in source languages and evaluating the resulting models across different query languages~\cite{xiangmultilingual,farashah2026multilingual}. 
Moving beyond independent per-language evaluation, Hwang et
al.~\cite{hwang2026knowledge} focus on the consistency of knowledge removal across
language pairs. 
These studies reveal whether forgetting transfers across semantically equivalent multilingual prompts. However, transfer does not always  match the intended objective: for globally scoped requests, the target information should become inaccessible across all languages; for language-conditioned requests, forgetting should remain confined to the designated language while access in other languages is preserved. We term these the \emph{common-goal} and \emph{language-conditioned} settings, respectively. Existing multilingual evaluations do not explicitly distinguish between these objectives and therefore cannot determine whether forgetting remains within its intended boundary.

\input{tables/benchmark_description}
In this paper, we propose \textbf{\shortname{}} (Cross-Lingual and Language-Bound Protocol for LLM Unlearning), a multilingual benchmark with three key designs. First, \shortname{} defines two settings: {common-goal forgetting}, where the forgetting effect should propagate across languages, and {language-conditioned forgetting}, where it should remain confined to a specific language context. Second, \shortname{} adopts a hierarchical two-stage matching scheme. At the topic level, goal-guided target--neighbor pairing places semantically comparable topics on opposite sides of the intended forgetting boundary; at the level of knowledge units, {schema-aware relation matching} aligns these units by relation type, semantic slot, answer type, and supporting evidence, yielding one-to-one forget--retain units with closely matched semantics. Third, each knowledge unit is first instantiated as an English QA family consisting of a core QA and semantically equivalent variants, and the QA family is translated into the other nine languages using dual-anchor multilingual translation.
We also construct core QAs based on holdout units for membership inference attacks. 
Finally, \shortname{} contains 800 knowledge-unit pairs and 72,000 multilingual QA instances.  

Extensive experiments with six representative LLM unlearning methods reveal that the central challenge in multilingual unlearning is not merely whether forgetting transfers, but whether its propagation can be controlled. Across the six representative methods, we observe opposite failure modes under the two objectives: cross-lingual forgetting remains incomplete when common-goal forgetting requires universal suppression, yet propagates excessively when language-conditioned forgetting requires confinement to the source language. General multilingual utility is also an unreliable proxy for selective preservation, as a model may retain near-original performance on a general benchmark while substantially damaging relation-matched non-target knowledge. Together, these findings call for multilingual unlearning methods that jointly account for propagation control and local knowledge preservation. Our main contributions are summarized as follows: 
\begin{itemize}[leftmargin=1.0em]
    \item We formulate multilingual LLM unlearning as a {propagation-boundary} problem and define two settings: common-goal forgetting and language-conditioned forgetting.
    
    \item We introduce \shortname{}, a multilingual LLM unlearning benchmark featuring complementary forgetting objectives, hierarchical matching at the topic and knowledge-unit levels, and semantically equivalent variants. 
    
    \item Experiments with six representative methods on Llama-3.1-8B-Instruct show that current methods fail in opposite directions: forgetting remains incomplete under the common-goal setting, yet propagates beyond the intended scope under the language-conditioned setting.
\end{itemize}

%% file: tex/2related_work.tex
\section{Related Work}

\subsection{LLM Unlearning Benchmarks}

LLM unlearning benchmarks assess whether a method can suppress designated knowledge while preserving non-target capabilities~\cite{yoon2025r,dorna2026openunlearning}. Text-based benchmarks cover increasingly diverse knowledge and evaluation settings. TOFU evaluates forgetting and retention over fictitious author profiles~\cite{maini2024tofu}, whereas RWKU extends evaluation to real-world entity knowledge and introduces diverse probes for detecting residual knowledge~\cite{cao2024rwku}. MUSE further evaluates memorization, privacy leakage, utility preservation, scalability, and sequential unlearning requests~\cite{shi2025muse}, while WMDP focuses on hazardous knowledge in biosecurity, cybersecurity, and chemical security~\cite{li2024wmdp}. Recent studies have examined preservation at a finer granularity. Chang et al. analyze how retain-set composition affects unlearning and show that syntactically similar neighboring queries are particularly vulnerable to performance degradation~\cite{chang2025retain}. DUSK evaluates whether unlearning methods can remove forget-specific knowledge while preserving knowledge shared between the forget and retain data~\cite{jeung2026dusk}. Recent work has also extended unlearning benchmarks to multimodal LLMs, including fictitious facial identities and multimodal misinformation~\cite{ma2025benchmarking,zheng2026offside,wang2026icu}. 

Despite this progress, existing benchmarks generally do not evaluate whether forgetting propagates according to an intended linguistic scope. Moreover, preservation is commonly measured using broad utility tasks or retain knowledge that is not explicitly matched to the forget targets. Such evaluations may therefore overlook localized damage to semantically neighboring knowledge and cannot distinguish insufficient cross-lingual propagation from propagation beyond the intended boundary.

\subsection{Multilingual LLM Unlearning}

Multilingual LLM unlearning examines how an intervention applied in one or more source languages affects access to the same underlying knowledge across different query languages~\cite{xiangmultilingual,farashah2026multilingual}. Existing studies develop language-aware unlearning strategies, investigate misinformation removal across languages, and evaluate source--target language configurations~\cite{choi2024cross,lu2025learn}. Their results show that cross-lingual forgetting is often incomplete, uneven, and asymmetric, with transfer patterns influenced by linguistic characteristics and shared multilingual representations~\cite{lizzo2026evaluating,farashah2026multilingual,chen2026emergence}.

% However, existing studies mainly characterize whether forgetting transfers across languages, without explicitly distinguishing the common-goal and language-conditioned objectives, and relation-matched knowledge preservation. 

However, existing studies mainly characterize whether forgetting transfers across languages, without explicitly distinguishing between the common-goal and language-conditioned objectives or evaluating the preservation of relation-matched knowledge.

%% file: tex/3construction_simplified.tex
\section{Benchmark Construction}
\label{sec:benchmark_construction}

\begin{figure*}[t]
    \centering
    \includegraphics[width=0.82\textwidth]{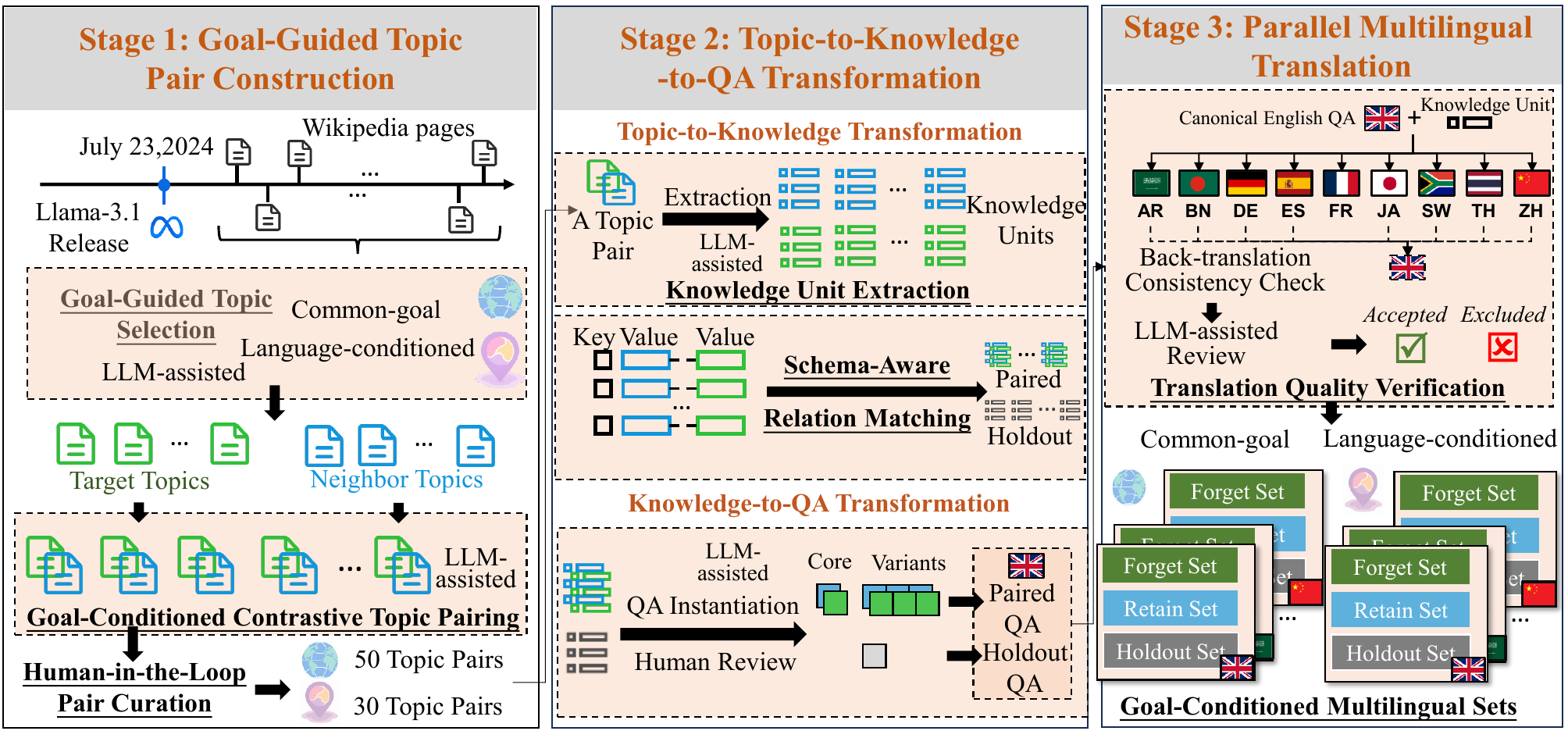}
    \caption{Overview of the \shortname\ construction pipeline.}
    \label{fig:overall}
\end{figure*}

In this section, we introduce \shortname{}, a \fullname{} benchmark. As illustrated in Figure~\ref{fig:overall}, its construction consists of Goal-Guided Topic Pair Construction, Topic-to-Knowledge-to-QA Transformation, and Dual-Anchor Multilingual Translation.

\subsection{Goal-Guided Topic Pair Construction}
\label{sec:topic_construction}

To define the semantic boundary of forgetting at the topic level, we construct topic pairs consisting of a \emph{target topic} and a \emph{neighbor topic}. The target topic represents knowledge within the intended forgetting boundary, whereas the neighbor topic represents semantically related knowledge outside that boundary. 

\subsubsection{Goal-Guided Topic Selection}

In \shortname{}, each topic corresponds to an English Wikipedia page. To reduce the likelihood that the target knowledge has already been encountered during pretraining, we consider only topics whose defining event, discovery, product release, legal change, or emergence in public discourse postdates the release of Llama~3.1, the base LLM used in our experiments. Candidate topics must also contain sufficient verifiable factual information to support subsequent knowledge-unit extraction and relation matching.

Topic selection is further guided by the two forgetting settings. We denote the language used to perform unlearning as the source language ($s$), and the language used to query the unlearned model as the evaluation language ($t$). In the {common-goal} setting, the intended forgetting boundary spans all evaluation languages rather than depending on a particular language context. Therefore, regardless of the source language ($s$) in which unlearning is performed, the target knowledge should become inaccessible in every evaluation language ($t$). This setting thus requires the forgetting effect to propagate across languages, achieving universal suppression of the target knowledge. In the {language-conditioned} setting, each target topic is assigned a source language ($s$), and the forgetting effect is expected to remain confined to that language context. When the evaluation language matches the source language, i.e., ($t=s$), the target knowledge should be suppressed; when ($t\neq s$), it should remain accessible in the other languages. The two settings serve different evaluation purposes. The common-goal setting follows the conventional semantics of machine unlearning~\cite{han2025trustworthy,bourtoule2021machine}, under which the target knowledge should become inaccessible regardless of the language used to query the model. The language-conditioned setting, by contrast, does not necessarily correspond to a unique data-deletion or retraining oracle. Instead, it operationalizes a benchmark-specific conditional-access objective, in which the behavioral effect of an intervention is expected to remain within a designated language boundary. We include this setting as a stress test of propagation control in multilingual models, rather than as a claim that the underlying knowledge has been completely removed from the model.

\subsubsection{Contrastive Topic Pairing}

For each target topic, we use a Codex-assisted workflow to identify a semantically related neighbor topic. A valid pair should belong to the same semantic domain, have comparable levels of abstraction and temporal contexts, and contain similar types of facts, while referring to distinct entities or events with different factual answers. We exclude parent--child topics, alternative formulations or variants of the same event, and superficial matches. 

% In the {common-goal} setting, the neighbor topic serves as a semantically adjacent reference that remains outside the intended forgetting boundary. For example, the 2024 Shenzhen stabbing can be paired with the 2024 Wuxi stabbing. Both are structurally similar public-safety incidents and contain comparable facts about dates, locations, casualties, and official responses, while involving distinct events with non-overlapping factual answers.

In the {common-goal} setting, the neighbor topic serves as a semantically adjacent reference that remains outside the intended forgetting boundary. For example, Hurricane Helene can be paired with Hurricane Milton. Both are major Atlantic hurricanes from the 2024 season and contain comparable facts about affected areas, casualties, and damage, while representing distinct storms with non-overlapping factual answers.

In the {language-conditioned} setting, the neighbor topic must additionally remain accessible in the source language ($s$), enabling a direct comparison between target suppression and nearby knowledge preservation within the same language context. For example, the opening and closing ceremonies of the 2024 Summer Olympics form a suitable pair: they belong to the same event domain and contain comparable facts about dates, venues, performances, and public responses, while describing distinct events. When the opening ceremony is designated for forgetting in a source language, the closing ceremony remains outside the intended forgetting boundary and should be preserved in all evaluation languages, including the source language. Because inaccurate language assignments or poorly matched neighbors may blur the intended forgetting boundary, we prioritize boundary precision over broader topic coverage in the {language-conditioned} setting. 
% Detailed selection criteria and human-review procedures are provided in Appendix~\ref{app:stage1_details}.

\subsection{Topic-to-Knowledge-to-QA Transformation}
\label{sec:topic_to_qa}

\subsubsection{Topic-to-Knowledge Transformation}
\label{sec:transform1}

Topic-level comparability does not imply fact-level comparability: two related topics may describe different attributes or support different answer types. We therefore decompose each target--neighbor topic pair into \emph{knowledge units}, where each unit is an atomic fact with a concise answer and directly traceable evidence. A unit records its relation type, semantic slot, answer and answer type, factual statement, and supporting source span. Gemini~3.1 Pro Preview \cite{team2026gemini} first identifies relations supported by both topics and then extracts evidence-grounded candidate units under this shared relation inventory. This representation makes the evaluated knowledge explicit and separates it from any particular question wording.

We then apply {schema-aware relation matching} to construct fact-level contrasts across the forgetting boundary. A target-topic unit and a neighbor-topic unit are eligible only when they have the same normalized relation type, compatible semantic slots and answer types, and direct evidence in their respective sources. Eligible pairs are ranked by schema consistency and evidence quality, followed by one-to-one selection without unit reuse. The target-side member of each match becomes a \emph{forget unit}, and the neighbor-side member becomes its \emph{retain unit}; together they form a relation-aligned forget--retain probe. Unmatched valid units are sampled separately as \emph{holdout units} for membership-inference evaluation, ensuring knowledge-unit-level disjointness from the matched pool. We manually review all selected units and matches. The complete schema, extraction filters, coverage-aware selection rule, holdout construction, and review criteria are given in Appendix~\ref{app:stage2_details}.

\subsubsection{Knowledge-to-QA Transformation}
\label{sec:transform2}

Each selected knowledge unit is next instantiated as a canonical English QA family. Gemini~3.1 Pro Preview receives only the unit's structured fields, rather than the full topic article, and jointly processes each matched forget--retain pair to preserve their fact-level comparability. For every forget or retain unit, it generates one \emph{core QA} that directly queries the encoded fact and three \emph{surface variants} that express the same information need with different wording. Across all four realizations, the relation type, semantic slot, expected answer, answer boundary, and supporting evidence must remain invariant.

This core-and-surface design distinguishes prompt sensitivity from a change in knowledge accessibility: failure on one wording is weak evidence of forgetting, whereas consistent degradation across equivalent realizations more directly reflects reduced access to the underlying fact. Because holdout units serve only as non-members in membership-inference attacks, each is converted into a single core QA without surface variants. Codex powered by GPT-5.5~\cite{barman2026gpt} independently reviews the generated QAs, after which human reviewers resolve invalid or ambiguous cases. Appendix~\ref{app:stage2_details} specifies the answer-anchored generation protocol, admissible surface variation, and full review procedure.

\subsection{Parallel Multilingual QA Construction}
\label{sec:parallel_multilingual_translation}

\subsubsection{Dual-Anchor Translation}

\shortname{} covers ten benchmark languages: Arabic
(\texttt{ar}), Bengali (\texttt{bn}), German (\texttt{de}),
English (\texttt{en}), Spanish (\texttt{es}), French
(\texttt{fr}), Japanese (\texttt{ja}), Swahili (\texttt{sw}),
Thai (\texttt{th}), and Chinese (\texttt{zh}). 
% We use the
% corresponding uppercase codes in figures and tables.

Reliable cross-lingual evaluation requires QAs in different languages to access the same underlying knowledge. However, unconstrained translation may alter the information being queried. For example, translating ``which city'' as a generic ``where'' question broadens the expected-answer boundary from a city to a country or venue. We therefore introduce \emph{Dual-Anchor Parallel QA Translation}, which conditions translation on both the structured knowledge unit and its English QA. The knowledge-unit anchor specifies what must be asked by fixing the underlying fact, relation type, semantic slot, and expected-answer boundary; the QA anchor specifies how it should be asked by preserving the question intent and whether the instance is a core QA or a surface variant. A valid translation must satisfy both anchors while changing only its linguistic realization. For each core QA or surface variant, Gemini~3.1 Pro Preview jointly receives the corresponding structured knowledge unit and English QA and produces a QA in each of the other nine benchmark languages. The complete translation procedure is provided in Appendix~\ref{app:stage3_details}.

\subsubsection{Translation Verification}
We employ a three-stage verification procedure. First, Gemini~3.1 Pro Preview blindly back-translates each multilingual QA into English without access to its original English QA~\cite{brislin1970back}. In a separate verification step, the back-translation is compared with the original for consistency in question intent, answer semantics, expected-answer boundary, and whether it is a core QA or a surface variant; failed instances are retranslated and verified again. Second, Codex powered by GPT-5.5 independently reviews every translated QA against both anchors. Third, we audit the reliability of these model-based decisions by conducting 20 rounds of human verification, each based on 100 randomly sampled QAs accepted by GPT-5.5. 
% Nearly all sampled QAs satisfy the dual-anchor constraints, indicating high agreement between GPT-5.5 and human reviewers; any errors identified during the audit are corrected.

The verified QAs are organized into three multilingual sets according to their underlying knowledge units. QAs derived from matched forget and retain units form the forget set {\small$\mathcal{Q}_f=\bigcup_{t\in\mathcal{L}}\mathcal{Q}_f^t$} and retain set {\small$\mathcal{Q}_r=\bigcup_{t\in\mathcal{L}}\mathcal{Q}_r^t$}, respectively, where $\mathcal{L}$ denotes the set of ten benchmark languages and $t$ is the evaluation language. QAs derived from holdout units form the holdout set {\small$\mathcal{Q}_h=\bigcup_{t\in\mathcal{L}}\mathcal{Q}_h^t$}. Each forget or retain unit contributes one core QA and three semantically equivalent surface variants per language, whereas each holdout unit contributes only one core QA per language. Further implementation details  are provided in Appendix~\ref{app:stage3_details}.

%% file: tex/3method.tex
\section{Evaluation Protocol}
\label{sec:evaluation_protocol}

\subsection{LLM Unlearning Pipeline}
\label{sec:unlearning_pipeline}

Our evaluation starts with multilingual QA injection. Specifically, we perform full-parameter supervised fine-tuning (SFT) on the base LLM, Meta-Llama-3.1-8B-Instruct~\cite{dubey2024llama}, using the core QAs from the forget and retain sets in all ten benchmark languages. The resulting Original model serves as the common starting point for all unlearning methods. We then perform LLM unlearning separately for each source language. 
% Let $\mathcal{L}$ denote the ten languages. 
% For each source language {\small$s\in\mathcal{L}$}, we apply an unlearning method to the Original model using only the corresponding forget and retain core QAs in  {\small$s$}, producing a source-specific unlearned model {\small$M_{\mathrm{un}}^{(s)}$}. We then evaluate {\small$M_{\mathrm{un}}^{(s)}$} in every evaluation language {\small$t\in\mathcal{L}$}.
Recall that $\mathcal{L}$ denotes the set of ten benchmark languages. For each source language {\small$s\in\mathcal{L}$}, we apply an unlearning method to the Original model using only the corresponding forget and retain core QAs in {\small$s$}, producing a source-specific unlearned model {\small$M_{\mathrm{un}}^{(s)}$}. We then evaluate {\small$M_{\mathrm{un}}^{(s)}$} in every evaluation language {\small$t\in\mathcal{L}$}.
Performance at {\small$t=s$} characterizes source-language behavior, whereas performance at {\small$t\neq s$} characterizes the cross-lingual effects of unlearning.

For the {common-goal} setting, we additionally construct a Retrain model following the standard retraining principle in machine unlearning~\cite{bourtoule2021machine}. We perform the same full-parameter SFT on the base LLM using only the retain set, which provides a reference for model behavior without the target knowledge. We omit this reference for the {language-conditioned} setting: retaining the forget set in non-source languages may transfer it back to the source language, whereas removing it globally would violate the specific preservation objective in this setting.

\subsection{Multilingual Evaluation}

\subsubsection{Knowledge-Accessibility Metrics}
For each source--evaluation-language pair {\small$(s,t)\in\mathcal{L}^2$}, we evaluate {\small$M_{\mathrm{un}}^{(s)}$} using the corresponding forget and retain QAs realized in evaluation language {\small$t$}. For each QA, we compare the model-generated answer with its expected answer using four answer-level measures: 1) Exact Match (EM), 2) ROUGE-L (RL)~\cite{lin2004rouge}, 3) Sentence Similarity (SS), and 4) LLM-as-a-Judge. EM measures exact equality after output normalization, while ROUGE-L computes the longest-common-subsequence-based F1 score. SS is the cosine similarity between the L2-normalized BGE-M3 dense embeddings of the generated and expected answers~\cite{chen2024m3embedding}. LLM-as-a-Judge evaluates answer correctness using an evaluator LLM; we exclude it from the main comparison because evaluator- and language-dependent calibration may confound the cross-lingual effects under study.

For each answer-level measure {\small$m$}, we first average the scores of the core QA and three surface variants corresponding to each knowledge unit and then average across knowledge units, yielding the \emph{Knowledge Accessibility} score {\small$A_z^m(s,t)$}. Here, {\small$z\in\{f,r\}$} indexes the forget and retain roles, and {\small$m\in\{\mathrm{EM},\mathrm{RL},\mathrm{SS},\mathrm{LLM\text{-}as\text{-}a\text{-}Judge}\}$} denotes the answer-level measure. We summarize the resulting language-pair matrix using two macro-averaged scores:
{\small$
\hat{A}_z^m(\mathrm{S})
=
\frac{1}{|\mathcal{L}|}
\sum_{s\in\mathcal{L}}
A_z^m(s,s),
$}
{\small$
\hat{A}_z^m(\mathrm{C})
=
\frac{1}{|\mathcal{L}|}
\sum_{s\in\mathcal{L}}
\left[
\frac{1}{|\mathcal{L}|-1}
\sum_{t\in\mathcal{L}\setminus\{s\}}
A_z^m(s,t)
\right].
$} 
Here, lowercase {\small$s$} and {\small$t$} index individual source and evaluation languages, respectively. The quantities {\small$\hat{A}_z^m(\mathrm{S})$} and {\small$\hat{A}_z^m(\mathrm{C})$} are the macro-averaged Source and Cross results over the ten diagonal and ninety off-diagonal language-pair cells, respectively.

In the {common-goal} setting, successful unlearning requires both {\small$\hat{A}_f^m(\mathrm{S})$} and {\small$\hat{A}_f^m(\mathrm{C})$} to be low because the target knowledge should become inaccessible in every evaluation language. In the {language-conditioned} setting, {\small$\hat{A}_f^m(\mathrm{S})$} should be low, whereas {\small$\hat{A}_f^m(\mathrm{C})$} should remain high because the same target knowledge should remain accessible in non-source languages. In both settings, accessibility to the matched retain knowledge derived from neighbor topics, measured by {\small$\hat{A}_r^m(\mathrm{S})$} and {\small$\hat{A}_r^m(\mathrm{C})$}, should remain high.

\subsubsection{Other Metrics}
We further evaluate Membership Inference Attacks (MIA) using Loss, Min-K\%~\cite{shi2024detecting}, Min-K\%++~\cite{zhang2025minkpp}, and Zlib~\cite{carlini2021extracting}. For each source-specific unlearned model {\small$M_{\mathrm{un}}^{(s)}$}, the forget core QAs used during knowledge injection are treated as members, while the unseen holdout core QAs in the same source language {\small$s$} are treated as non-members. Each attack is evaluated independently for every source language using raw ROC-AUC, after which we macro-average across the ten source-language runs. An AUC of {\small$0.5$} indicates chance-level discrimination; values below {\small$0.5$} may reflect reversed attack-score ordering rather than stronger unlearning.

We additionally evaluate general multilingual utility using Belebele~\cite{bandarkar2024belebele}. Each source-specific unlearned model is evaluated zero-shot on 900 multiple-choice examples in each of the ten benchmark languages. We first average accuracy across the ten evaluation languages for each source-specific model and then macro-average across the ten source-language runs, yielding one overall Belebele accuracy for each method under each setting. Higher accuracy indicates better preservation of general multilingual utility. Further details are in Appendix~\ref{app:eval}, e.g., metric formulations.

%% file: tex/4experiment.tex
\input{tables/main_results}
\section{Experiments}
In this section, we address the following research questions (RQs): 
\begin{itemize}[leftmargin=1.0em]
    \item \textbf{RQ1:} How do existing methods perform under two objectives? 
    \item \textbf{RQ2:} What are the boundaries of their cross-lingual transfer?
    \item \textbf{RQ3:} How do they perform according to other metrics? 
    \item \textbf{RQ4:} Do surface variants yield results consistent with those of core QAs?
    \item \textbf{RQ5:} How does forced response-language switching affect knowledge accessibility?
% `RQ1: How do existing methods perform under the two forgetting objectives?`；`RQ2: What are the boundaries of their cross-lingual transfer?`；`RQ3: How do they perform according to complementary metrics?`；`RQ4: Do surface variants yield results consistent with those of core QAs?`；`RQ5: How does forced response-language switching affect knowledge accessibility?`
\end{itemize}
Due to page limits, more experiments can be found in Appendix \ref{app:exp}.

\input{tables/culture_results}

\subsection{Experimental Settings}
\subsubsection{Models and Data}
During both knowledge injection and unlearning, we use only core QAs and reserve variants for robustness evaluation. The Retrain model is initialized from the same base LLM and trained on the retain set. We select Llama-3.1-8B-Instruct as the target LLM for two reasons. First, the defining events of all selected topics postdate its release, reducing the likelihood that the evaluated knowledge was encountered during pretraining. Second, CLLPU follows the task-specific target-LLM protocol adopted by existing unlearning benchmarks~\cite{shi2025muse,jeung2026dusk,ramakrishna2025semeval}. For example, MUSE assigns a designated target LLM to each corpus~\cite{shi2025muse}, while DUSK instantiates its benchmark with Llama-3-8B~\cite{jeung2026dusk}. Our objective is to introduce and validate a new evaluation dimension—whether forgetting remains within its intended linguistic boundary—rather than to claim model-invariant propagation patterns.

\subsubsection{Compared Methods}
We evaluate six representative LLM unlearning methods under the same source-language data protocol: GA and GD~\cite{yao2024large}, NPO~\cite{zhang2024negative}, SimNPO~\cite{fan2025simplicity}, BalDRO-NPO, and BalDRO-SimNPO~\cite{shao2026baldro}. In addition, Original and Retrain are included only as reference models.

\subsubsection{Implementation Details}
To obtain the Original and Retrain models, we use a learning rate of {\small$1\times10^{-5}$}, the AdamW optimizer with a weight decay of {\small$0.01$}, linear warmup for one epoch, a per-device batch size of {\small$8$}, four gradient-accumulation steps, and bfloat16 precision. As for the unlearning process, we use AdamW and select the learning rate from {\small$\{1\times10^{-5},3\times10^{-5},5\times10^{-5}\}$}. The effective batch size is {\small$32$} for common-goal runs and {\small$8$} for language-conditioned runs. More details about settings are in Appendix~\ref{app:exp}.

\subsection{Overall Performance (RQ1)}

\begin{figure*}[t]
    \centering
    \includegraphics[width=\textwidth]{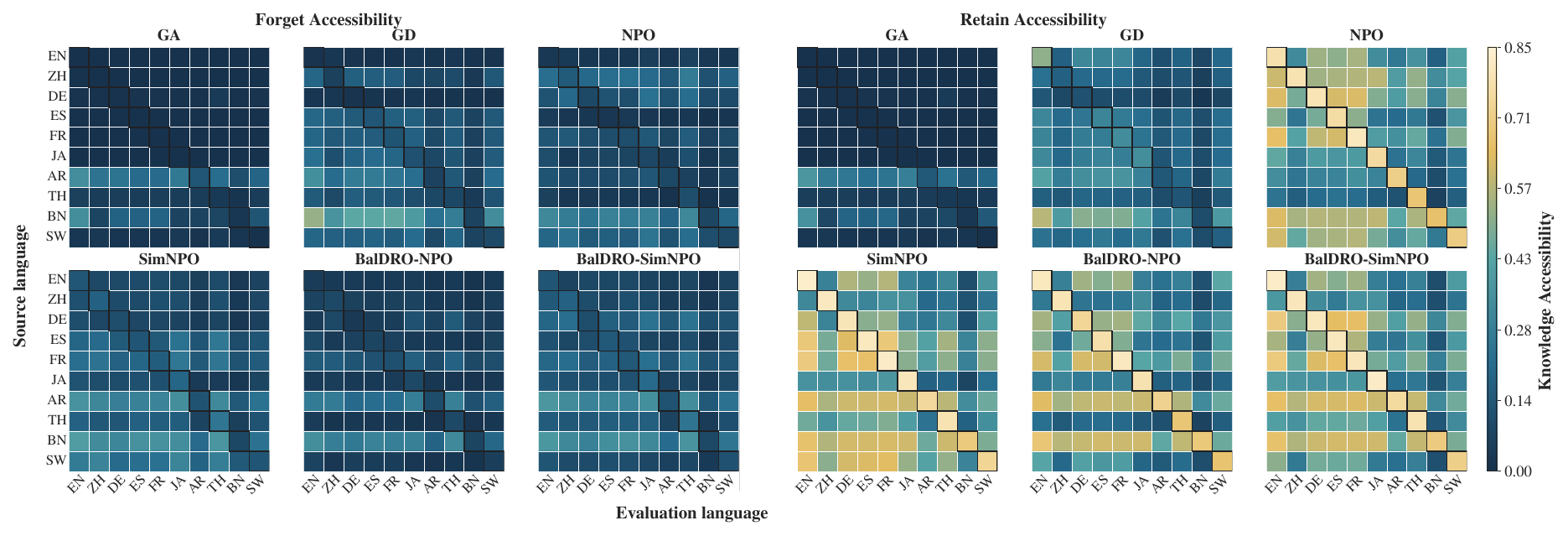}
        \caption{{ROUGE-L in the common-goal setting.}
Diagonal cells report source accessibility {\small$A_z^{\mathrm{RL}}(s,s)$}, while off-diagonal cells report cross accessibility {\small$A_z^{\mathrm{RL}}(s,t)$}, where {\small$t\neq s$}. Lower forget accessibility and higher retain accessibility indicate better performance.}
    \label{fig:RL_com}
\end{figure*} 
Tables~\ref{tab:common_goal_overall} and~\ref{tab:culture_specific_overall} report the overall knowledge accessibility results. We summarize four main findings:
\begin{itemize}[leftmargin=1.2em, itemsep=0.35em, topsep=0.3em]

\item \textbf{Common-goal forgetting transfers across languages, but incompletely.}
All methods reduce Cross forget accessibility relative to the Original model, showing that unlearning in one source language affects the same knowledge in other languages. However, Cross accessibility remains consistently higher than Source accessibility, indicating that the forgetting effect weakens during cross-lingual transfer.

\item \textbf{None of the six evaluated methods establishes the desired language-conditioned boundary.}
Reducing Source accessibility also substantially reduces accessibility in non-source languages. For example, GD achieves a Cross ROUGE-L of only 0.1788, compared with 0.8068 for the Original model. Similarly, the highest Cross SS among all methods is only 0.5103, far below the Original result of 0.8910. Current methods therefore cannot reliably forget knowledge only in the designated source language.

\item \textbf{Stronger forgetting often comes at the cost of retain Accessibility.}
GA achieves the lowest forget accessibility, but its Source ROUGE-L scores on the forget and retain sets are similarly low at 0.0192 and 0.0201. This result indicates that its strong forgetting is accompanied by broad damage to related knowledge rather than selective removal of the target knowledge.

\item \textbf{Embedding similarity can overestimate residual knowledge accessibility.} In the common-goal setting, GA achieves a Source EM of 0.0070 but a Source SS of 0.3273. This discrepancy should not be interpreted simply as evidence that substantial target knowledge remains accessible. Because our forget and retain units are relation-matched, their answers can be semantically similar despite being factually distinct; indeed, the BGE-M3 similarity between paired forget--retain answers is already approximately 0.48--0.50. Therefore, SS need not approach zero before the target fact becomes inaccessible: a score around 0.5 may already correspond to a plausible but factually incorrect response that is semantically close to neighboring knowledge.  Details can be found in Appendix \ref{app:exp}. 

\end{itemize}

\begin{figure*}[t]
    \centering
    \includegraphics[width=\textwidth]{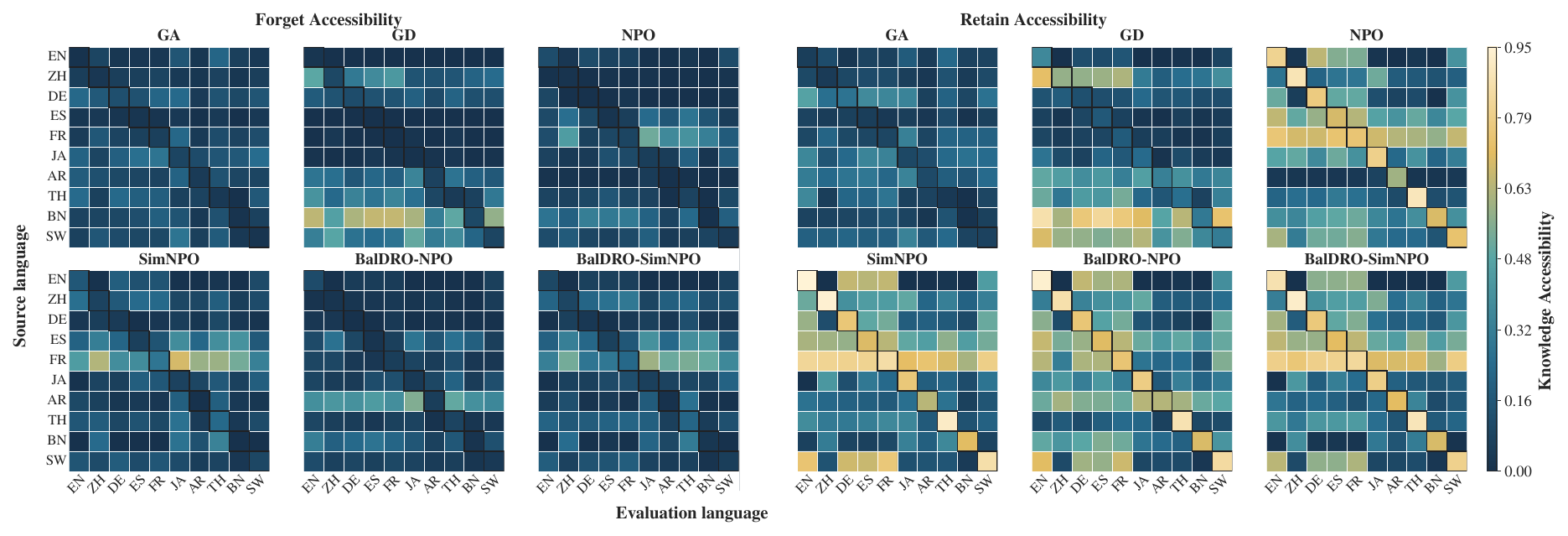}
    \caption{{ROUGE-L in the language-conditioned setting.}
    Diagonal cells report source accessibility {\small$A_z^{\mathrm{RL}}(s,s)$}, while off-diagonal cells report cross accessibility {\small$A_z^{\mathrm{RL}}(s,t)$}, where {\small$t\neq s$}. Successful unlearning requires low forget accessibility on the diagonal, high forget accessibility off the diagonal, and high retain accessibility across all languages.}
    \label{fig:RL_cul}
\end{figure*}
\input{tables/othermetrics}
\subsection{Cross-Lingual Transfer Boundaries (RQ2)}
\label{sec:transfer_boundaries}
Figures~\ref{fig:RL_com} and~\ref{fig:RL_cul} show pairwise ROUGE-L accessibility; the corresponding EM and SS heatmaps are provided in Appendix~\ref{app:heat}. We make four observations. \textbf{First}, common-goal forgetting transfers across languages, but its strength depends strongly on the source language. With GA, using English or Japanese as the source reduces target accessibility to nearly zero in all evaluation languages. In contrast, using Arabic as the source yields an average cross accessibility of 0.221. This variation shows that the choice of source language can substantially change the cross-lingual effect of the same unlearning method. \textbf{Second}, cross-lingual forgetting is directional rather than symmetric. Under GD, target accessibility from Bengali to English is 0.519, while the reverse direction is only 0.009. Thus, the effect from language {\small$s$} to language {\small$t$} cannot be inferred from the effect from {\small$t$} to {\small$s$}; each direction defines a different unlearning boundary. \textbf{Third}, preserving utility in the source language does not guarantee preservation in other languages. For example, in the language-conditioned setting, NPO with Arabic as the source language achieves a retain accessibility of 0.596 in Arabic but only 0.030 on average in other languages. \textbf{Finally}, strong forgetting is often not selective. GA produces similar low-value patterns for Forget Accessibility and Retain Accessibility. Its strong forgetting therefore reflects broad knowledge damage rather than precise removal of the target knowledge.

\subsection{Other Metrics (RQ3)}

% 我们还考虑了LLM-as-judge
We draw four findings from Table~\ref{tab:other_metrics}. \textbf{First}, the {common-goal} setting shows a trade-off between membership privacy and general utility. GA achieves AUCs closest to 0.5 but lowers Belebele accuracy to 0.5597, whereas GD preserves the highest accuracy of 0.7186 but has AUCs farther from 0.5. \textbf{Second}, the six evaluated methods still perform poorly in the language-conditioned setting. Even the best method, SimNPO, produces AUCs of 0.3003--0.3869 rather than approaching 0.5. Together with Table~\ref{tab:culture_specific_overall}, this result shows the need for methods that can limit forgetting to the source language. \textbf{Third}, the reference models support the validity of our MIA evaluation. Original produces consistently high AUCs of 0.93--0.97, while Retrain approaches the random baseline of 0.5. \textbf{Finally}, Belebele and knowledge accessibility measure different aspects of utility. In the language-conditioned setting, GA maintains a Belebele accuracy of 0.7868, close to 0.8086 for Original, but severely damages the matched retain knowledge in Table~\ref{tab:culture_specific_overall}. This result shows that our matched-pair construction can detect damage to related knowledge that a general multilingual benchmark may miss. 

\subsection{Core QAs and Surface Variants (RQ4)}
\begin{figure*}[t]
    \centering
    \includegraphics[width=0.82\textwidth]{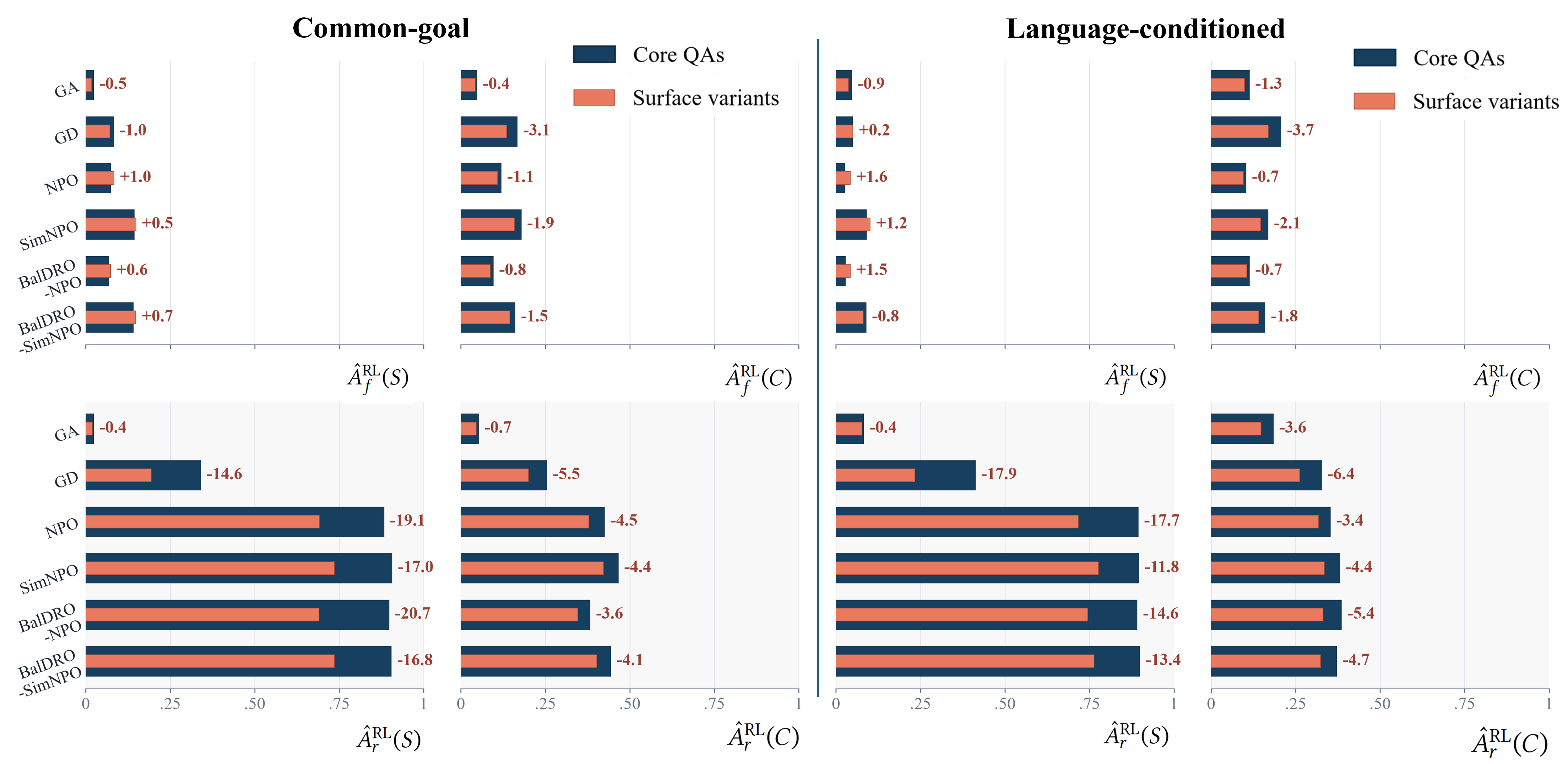}
    \caption{Performance on core QAs versus surface variants. Each surface-variant bar reports the mean ROUGE-L score across three variants. Values adjacent to the bars denote the surface-variant score minus the core-QA score in percentage points. }
    \label{fig:core_surface_robustness}
\end{figure*}

Figure~\ref{fig:core_surface_robustness} shows that forgetting generalizes well from core QAs to unseen surface variants: across methods, settings, and evaluation scopes, forget-set gaps remain between $-3.7$ and $+1.6$ percentage points, with predominantly negative Cross gaps indicating that variants rarely recover forgotten knowledge. Retain accessibility is more sensitive to wording, with Source gaps reaching $20.7$ points and smaller but non-negligible Cross gaps. Thus, core-only evaluation can characterize forgetting reasonably well but may overestimate retained knowledge accessibility. 

% 这四个，但其实计算的时候每个单元呗修改了就是{\small$A_z^{\mathrm{RL}}(s,t)$} 被改了。

\subsection{Forced Response-Language Switching(RQ5)}

\begin{figure}[t]
    \centering
    \includegraphics[width=0.85\columnwidth]{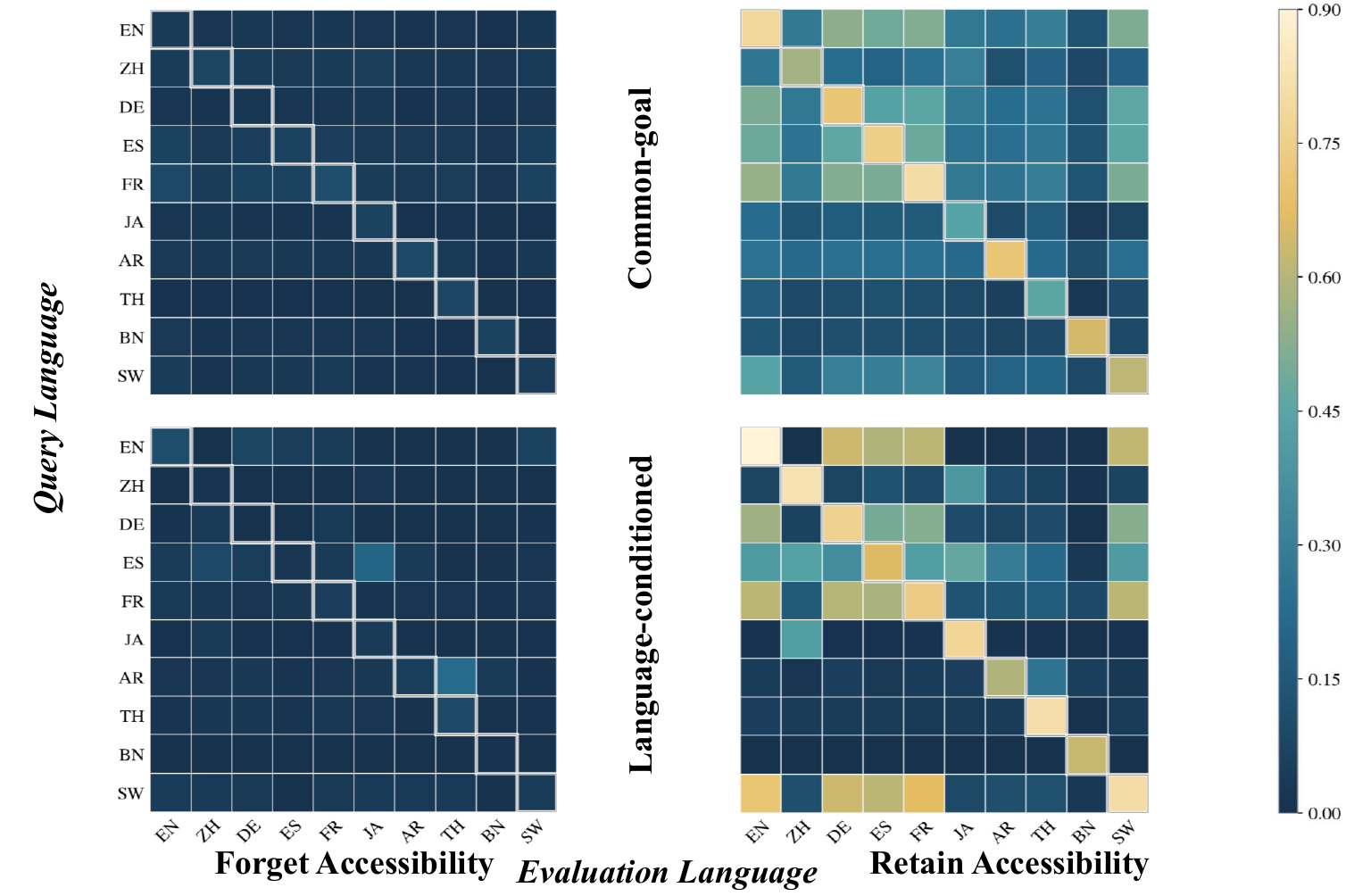}
    \caption{ROUGE-L accessibility under forced response-language switching. Given a query in language $s$, we instruct the model to respond in a designated language $u$.}
    \label{fig:forced_response_language}
    \vspace{-8mm}
\end{figure}

Figure~\ref{fig:forced_response_language} compares our main protocol against a constrained setting that keeps the query language fixed while forcing the model to respond in another language. Forced switching consistently reduces residual knowledge accessibility: the average forget-set ROUGE-L accessibility decreases from $0.0658$ on the diagonal to $0.0237$ off the diagonal in the common-goal setting, and from $0.0418$ to $0.0198$ in the language-conditioned setting. Thus, our main protocol constitutes the more stringent unlearning test: aligned-language evaluation recovers more residual knowledge and is therefore harder for an unlearning method to pass, whereas forced switching introduces an additional cross-lingual generation bottleneck that suppresses knowledge expression. Although a few language pairs, such as AR$\rightarrow$TH and ES$\rightarrow$JA, exhibit locally high accessibility, they do not alter the overall trend. The concurrent decrease in retain-set accessibility further indicates that the reduction partly arises from constrained generation rather than stronger forgetting. These results support direct multilingual evaluation as the primary protocol for exposing residual knowledge.

%% file: tables/main_results.tex
\begin{table*}[t]
\caption{\textbf{Overall performance in the common-goal setting.} Lower forget accessibility and higher retain accessibility indicate better performance. The best result is shown in \textbf{bold} and the second-best result is \underline{underlined}.}
\label{tab:common_goal_overall}

\centering
\resizebox{0.86\textwidth}{!}{
\begin{tabular}{l|cccccc|cccccc}
\hline

\multirow{2}{*}{
\diagbox{\textbf{Method}}{\textbf{Metric}}
}
&
\multicolumn{6}{c|}{\textbf{Forget Accessibility}}
&
\multicolumn{6}{c}{\textbf{Retain Accessibility}}
\\
\cline{2-7}
\cline{8-13}

&
{\small$\hat{A}_{f}^{\mathrm{EM}}(\mathrm{S})\downarrow$}
&
{\small$\hat{A}_{f}^{\mathrm{RL}}(\mathrm{S})\downarrow$}
&
{\small$\hat{A}_{f}^{\mathrm{SS}}(\mathrm{S})\downarrow$}
&
{\small$\hat{A}_{f}^{\mathrm{EM}}(\mathrm{C})\downarrow$}
&
{\small$\hat{A}_{f}^{\mathrm{RL}}(\mathrm{C})\downarrow$}
&
{\small$\hat{A}_{f}^{\mathrm{SS}}(\mathrm{C})\downarrow$}
&
{\small$\hat{A}_{r}^{\mathrm{EM}}(\mathrm{S})\uparrow$}
&
{\small$\hat{A}_{r}^{\mathrm{RL}}(\mathrm{S})\uparrow$}
&
{\small$\hat{A}_{r}^{\mathrm{SS}}(\mathrm{S})\uparrow$}
&
{\small$\hat{A}_{r}^{\mathrm{EM}}(\mathrm{C})\uparrow$}
&
{\small$\hat{A}_{r}^{\mathrm{RL}}(\mathrm{C})\uparrow$}
&
{\small$\hat{A}_{r}^{\mathrm{SS}}(\mathrm{C})\uparrow$}
\\
\hline

\textcolor{gray}{Original}
& \textcolor{gray}{0.6996}
& \textcolor{gray}{0.7660}
& \textcolor{gray}{0.8731}
& \textcolor{gray}{0.6996}
& \textcolor{gray}{0.7660}
& \textcolor{gray}{0.8731}
& \textcolor{gray}{0.7013}
& \textcolor{gray}{0.7630}
& \textcolor{gray}{0.8697}
& \textcolor{gray}{0.7013}
& \textcolor{gray}{0.7630}
& \textcolor{gray}{0.8697}
\\

\textcolor{gray}{Retrain}
& \textcolor{gray}{0.0735}
& \textcolor{gray}{0.2082}
& \textcolor{gray}{0.5675}
& \textcolor{gray}{0.0735} 
& \textcolor{gray}{0.2082} 
& \textcolor{gray}{0.5675}
& \textcolor{gray}{0.7005} 
& \textcolor{gray}{0.7626} 
& \textcolor{gray}{0.8686}
& \textcolor{gray}{0.7005} 
& \textcolor{gray}{0.7626} 
& \textcolor{gray}{0.8686}
\\
\hline

GA
& \underline{0.0070} & \textbf{0.0192} & \textbf{0.3273}
& \textbf{0.0123} & \textbf{0.0437} & \textbf{0.3397}
& 0.0070 & 0.0201 & 0.3288
& 0.0137 & 0.0467 & 0.3415
\\

GD
& 0.0197 & 0.0741 & \underline{0.3903}
& 0.0546 & 0.1435 & \underline{0.4431}
& 0.1365 & 0.2306 & 0.5489
& 0.1046 & 0.2136 & 0.5351
\\

NPO
& \textbf{0.0062} & 0.0810 & 0.4614
& 0.0326 & 0.1113 & 0.4962
& 0.6843 & 0.7387 & 0.8535
& 0.2832 & 0.3906 & 0.6959
\\

SimNPO
& 0.0314 & 0.1471 & 0.5275
& 0.0653 & 0.1638 & 0.5522
& \textbf{0.7296} & \textbf{0.7786} & \textbf{0.8769}
& \textbf{0.3238} & \textbf{0.4326} & \textbf{0.7273}
\\

BalDRO-NPO
& 0.0218 & \underline{0.0721} & 0.4400
& \underline{0.0319} & \underline{0.0895} & 0.4667
& 0.6909 & 0.7420 & 0.8541
& 0.2617 & 0.3551 & 0.6799
\\

BalDRO-SimNPO
& 0.0312 & 0.1463 & 0.5261
& 0.0579 & 0.1491 & 0.5447
& \underline{0.7286} & \underline{0.7776} & \underline{0.8761}
& \underline{0.3075} & \underline{0.4122} & \underline{0.7217}
\\
\hline

\end{tabular}
}
\end{table*}

%% file: tables/culture_results.tex
\begin{table*}[t]
\caption{{Overall performance in the language-conditioned setting.} Lower forget accessibility and higher retain accessibility indicate better performance. The best result is shown in \textbf{bold} and the second-best result is \underline{underlined}.}
\label{tab:culture_specific_overall}

\centering
\resizebox{0.86\textwidth}{!}{
\begin{tabular}{l|cccccc|cccccc}
\hline

\multirow{2}{*}{
\diagbox{\textbf{Method}}{\textbf{Metric}}
}
&
\multicolumn{6}{c|}{\textbf{Forget Accessibility}}
&
\multicolumn{6}{c}{\textbf{Retain Accessibility}}
\\
\cline{2-7}
\cline{8-13}

&
{\small$\hat{A}_{f}^{\mathrm{EM}}(\mathrm{S})\downarrow$}
&
{\small$\hat{A}_{f}^{\mathrm{RL}}(\mathrm{S})\downarrow$}
&
{\small$\hat{A}_{f}^{\mathrm{SS}}(\mathrm{S})\downarrow$}
&
{\small$\hat{A}_{f}^{\mathrm{EM}}(\mathrm{C})\uparrow$}
&
{\small$\hat{A}_{f}^{\mathrm{RL}}(\mathrm{C})\uparrow$}
&
{\small$\hat{A}_{f}^{\mathrm{SS}}(\mathrm{C})\uparrow$}
&
{\small$\hat{A}_{r}^{\mathrm{EM}}(\mathrm{S})\uparrow$}
&
{\small$\hat{A}_{r}^{\mathrm{RL}}(\mathrm{S})\uparrow$}
&
{\small$\hat{A}_{r}^{\mathrm{SS}}(\mathrm{S})\uparrow$}
&
{\small$\hat{A}_{r}^{\mathrm{EM}}(\mathrm{C})\uparrow$}
&
{\small$\hat{A}_{r}^{\mathrm{RL}}(\mathrm{C})\uparrow$}
&
{\small$\hat{A}_{r}^{\mathrm{SS}}(\mathrm{C})\uparrow$}
\\
\hline

\textcolor{gray}{Original}
& \textcolor{gray}{0.7915}
& \textcolor{gray}{0.8237}
& \textcolor{gray}{0.9047}
& \textcolor{gray}{0.7560}
& \textcolor{gray}{0.8068} 
& \textcolor{gray}{0.8910}
& \textcolor{gray}{0.7902} 
& \textcolor{gray}{0.8267} 
& \textcolor{gray}{0.9061}
& \textcolor{gray}{0.7459}
& \textcolor{gray}{0.8016}
& \textcolor{gray}{0.8895}
\\
\hline

GA
& \textbf{0.0116} & 0.0393 & \textbf{0.3812}
& 0.0238 & 0.1029 & 0.4237
& 0.0287 & 0.0783 & 0.4243
& 0.0525 & 0.1566 & 0.4723
\\

GD
& 0.0243 & 0.0499 & \underline{0.4261}
& \textbf{0.1163} & \textbf{0.1788} & 0.4857
& 0.2121 & 0.2780 & 0.5874
& 0.2062 & 0.2787 & 0.5720
\\

NPO
& \underline{0.0133} & \textbf{0.0381} & 0.4270
& 0.0520 & 0.0974 & 0.4583
& 0.7243 & 0.7611 & 0.8676
& 0.2492 & 0.3271 & 0.6629
\\

SimNPO
& 0.0458 & 0.0984 & 0.4888
& \underline{0.0924} & \underline{0.1523} & \textbf{0.5103}
& \textbf{0.7767} & \textbf{0.8067} & \textbf{0.8972}
& \underline{0.2578} & \textbf{0.3462} & 0.6686
\\

BalDRO-NPO
& \underline{0.0133} & \underline{0.0383} & 0.4271
& 0.0633 & 0.1074 & 0.4643
& 0.7525 & 0.7814 & 0.8789
& \textbf{0.2683} & \underline{0.3451} & \textbf{0.6789}
\\

BalDRO-SimNPO
& 0.0344 & 0.0824 & 0.4777
& 0.0877 & 0.1458 & \underline{0.5029}
& \underline{0.7675} & \underline{0.7976} & \underline{0.8920}
& 0.2550 & 0.3357 & \underline{0.6713}
\\
\hline

\end{tabular}
}
\end{table*}

%% file: tables/othermetrics.tex
\begin{table*}[t]
\centering
\caption{{Overall performance on membership inference and multilingual utility.} The best result is  shown in \textbf{bold} and the second-best result is \underline{underlined}.}
\label{tab:other_metrics}

\resizebox{0.84\textwidth}{!}{
\begin{tabular}{l|ccccc|ccccc}
\toprule

\multirow{3}{*}{\textbf{Method}}
&
\multicolumn{5}{c|}{\textbf{Common-goal}}
&
\multicolumn{5}{c}{\textbf{Language-conditioned}}
\\

\cmidrule(lr){2-6}
\cmidrule(lr){7-11}

&
\multicolumn{4}{c}{\textbf{Membership Inference (AUC {\small$\rightarrow 0.5$})}}
&
\multicolumn{1}{c|}{\textbf{Belebele} $\uparrow$}
&
\multicolumn{4}{c}{\textbf{Membership Inference (AUC {\small$\rightarrow 0.5$})}}
&
\multicolumn{1}{c}{\textbf{Belebele} $\uparrow$}
\\

\cmidrule(lr){2-5}
\cmidrule(lr){6-6}
\cmidrule(lr){7-10}
\cmidrule(lr){11-11}

&
\textbf{Loss}
&
\textbf{Min-K\%}
&
\textbf{Min-K\%++}
&
\textbf{Zlib}
&
\textbf{Accuracy}
&
\textbf{Loss}
&
\textbf{Min-K\%}
&
\textbf{Min-K\%++}
&
\textbf{Zlib}
&
\textbf{Accuracy}
\\

\midrule

{\color{gray}Original}
&
{\color{gray}0.9706}
&
{\color{gray}0.9683}
&
{\color{gray}0.9344}
&
{\color{gray}0.9673}
&
{\color{gray}0.8086}
&
{\color{gray}0.9727}
&
{\color{gray}0.9716}
&
{\color{gray}0.9393}
&
{\color{gray}0.9712}
&
{\color{gray}0.8086}
\\

{\color{gray}Retrain}
&
{\color{gray}0.4631}
&
{\color{gray}0.4669}
&
{\color{gray}0.4797}
&
{\color{gray}0.4402}
&
{\color{gray}0.8060}
&
{\color{gray}--}
&
{\color{gray}--}
&
{\color{gray}--}
&
{\color{gray}--}
&
{\color{gray}--}
\\

\midrule

GA
&
\textbf{0.5033}
&
\textbf{0.4927}
&
\textbf{0.5042}
&
\textbf{0.4307}
&
0.5597
&
0.1523
&
0.1449
&
0.2156
&
0.1767
&
\textbf{0.7868}
\\

GD
&
0.3246
&
0.3074
&
0.3340
&
0.3286
&
\textbf{0.7186}
&
0.1630
&
0.1488
&
0.1626
&
0.1671
&
0.6510
\\

NPO
&
0.1978
&
0.1924
&
0.2344
&
0.2274
&
0.5673
&
0.2154
&
0.1596
&
0.1937
&
0.1828
&
0.6365
\\

SimNPO
&
0.4160
&
0.3888
&
0.3673
&
0.3743
&
\underline{0.6107}
&
\textbf{0.3869}
&
\textbf{0.3103}
&
\textbf{0.3003}
&
\textbf{0.3103}
&
0.6512
\\

BalDRO-NPO
&
0.1633
&
0.1710
&
0.2020
&
0.2024
&
0.5309
&
0.1882
&
0.1453
&
0.1686
&
0.1670
&
\underline{0.6691}
\\

BalDRO-SimNPO
&
\underline{0.4178}
&
\underline{0.3933}
&
\underline{0.3789}
&
\underline{0.3761}
&
0.6055
&
\underline{0.3659}
&
\underline{0.2856}
&
\underline{0.2939}
&
\underline{0.2985}
&
0.6426
\\

\bottomrule
\end{tabular}
}
\end{table*}

%% file: tex/5conclusion.tex
\section{Conclusion and Future Work}

Reliable multilingual unlearning is essential for controlling how sensitive knowledge remains accessible across languages. A key challenge is that different requests require different propagation boundaries: some knowledge should be removed across all languages, whereas other knowledge should be forgotten only within a designated language context. To study this challenge, we formulated multilingual unlearning as a propagation-boundary problem and developed \shortname{}, a benchmark that jointly evaluates knowledge suppression, cross-lingual propagation, and selective preservation. Across the six methods evaluated on Llama-3.1-8B-Instruct, we observed opposite failure modes: forgetting remained incomplete when global removal was required, but propagated beyond the intended boundary when language-conditioned removal was required. In the future, we will investigate methods that can explicitly control cross-lingual unlearning propagation.

%% file: tex/6appendix.tex
\section{Benchmark Construction Details}
\label{app:benchmark_construction}

This appendix provides benchmark-construction details omitted from the main text. Additional resources related to the construction process—including the complete prompt templates, supporting Python code, and the human-in-the-loop workflow for interacting with Codex—are available in our repository: \url{https://github.com/CLLPU/CLLPU-bench}.

\begin{table*}[t]
\centering
\small
\setlength{\tabcolsep}{5pt}
\renewcommand{\arraystretch}{1.15}

\begin{tabularx}{
  0.8\textwidth
}{
  @{}
  l
  >{\raggedright\arraybackslash}X
  r
  r
  r
  @{}
}
\toprule
\textbf{Stage}
&
\textbf{Data Product}
&
\textbf{Common-goal}
&
\textbf{Language-conditioned}
&
\textbf{Total}
\\
\midrule

Stage 1
& Curated pairs of target-neighbor topics
& 50 & 30 & 80
\\

& Canonical Wikipedia pages
& 100 & 60 & 160
\\

\addlinespace[2pt]
Stage 2
& Extracted knowledge units
& 2,888 & 1,761 & 4,649
\\

& Selected relation-matched knowledge-unit pairs
& 500 & 300 & 800
\\

& Sampled holdout knowledge units
& 500 & 300 & 800
\\

& Canonical English QAs (core + variants)
& 4,000 & 2,400 & 6,400
\\

& Canonical English holdout QAs (core only)
& 500 & 300 & 800
\\

\addlinespace[2pt]
Stage 3
& Multilingual forget QAs (core + variants)
& 2,000 $\times$ 10 & 1,200 $\times$ 10 & 3,200 $\times$ 10
\\

& Multilingual retain QAs (core + variants)
& 2,000 $\times$ 10 & 1,200 $\times$ 10 & 3,200 $\times$ 10
\\

& Multilingual holdout QAs (core only)
& 500 $\times$ 10 & 300 $\times$ 10 & 800 $\times$ 10
\\

& {All multilingual QAs}
& {4,500 $\times$ 10}
& {2,700 $\times$ 10}
& {7,200 $\times$ 10}
\\

\bottomrule
\end{tabularx}

\caption{\textbf{Statistics of the benchmark construction pipeline.}
Each matched knowledge-unit pair contains one forget unit and one retain unit. Each unit is instantiated as one core QA and three semantically equivalent surface variants, whereas each holdout unit produces only one core QA. Multilingual counts include English and its nine translated versions. In the common-goal setting, all language realizations of forget knowledge are targeted; in the language-conditioned setting, only the designated source-language realization is targeted.}
\label{tab:construction_statistics}
\end{table*}

\subsection{Stage 1: Goal-Guided Topic Pair Construction}
\label{app:stage1_details}

\paragraph{Post-Release Boundary.}
The temporal cutoff used for topic selection is July~23, 2024, the release date of Llama~3.1~\cite{dubey2024llama}. A topic is considered post-cutoff only if its defining event, discovery, product release, legal change, or emergence in public discourse occurred after this date. For each candidate, we record the corresponding temporal evidence rather than relying solely on the creation or revision date of its Wikipedia page. Candidates with ambiguous temporal provenance or insufficient factual content are excluded.

\paragraph{Candidate Metadata.}
Each candidate topic is represented by an English Wikipedia page and annotated with its semantic domain, abstraction level, temporal context, cultural context, anticipated fact types, and the evidence supporting these attributes. These annotations support both goal-conditioned topic assignment and subsequent contrastive pairing. 
For each language-conditioned topic, we additionally record its designated source language, a rationale category (e.g., legal, cultural, public-safety, diplomatic, or public-discourse), and topic-specific evidence supporting the assignment. 
Assignments within each rationale category are reviewed by annotators with relevant linguistic, regional, or cultural familiarity. The reviewers assess the contextual relevance of the event, the likely scope of its legal, social, or cultural impact, and whether this scope provides a sufficiently clear basis for distinguishing one language context from the others. This evidence-backed process is intended to approximate plausible real-world motivations for language-bounded forgetting rather than relying on arbitrary topic--language mappings. Nevertheless, these rationales are used only to construct a contextually plausible, benchmark-specific source-language boundary; they should not be interpreted as legal determinations, universal cultural judgments, or claims that a topic has an intrinsic forgetting scope. Candidates with ambiguous impact boundaries or insufficiently supported rationales are excluded.

\paragraph{Contrastive Pairing Criteria.}
For each target topic, Codex powered by GPT-5.5 proposes a neighbor topic from a similar semantic domain and temporal context. The two topics must have comparable abstraction levels and contain parallel types of facts while referring to distinct entities or events. We exclude parent--child topics, alternative formulations or variants of the same event, and superficial matches obtained by substituting country or entity names. For language-conditioned pairs, the neighbor topic must additionally remain outside the forgetting scope associated with the source language.

\paragraph{Human Curation.}
All proposed target--neighbor topic pairs undergo full human review for semantic comparability, abstraction-level consistency, availability of parallel facts, and boundary separation. Each pair is reviewed by an English-proficient annotator with relevant cultural or regional familiarity, who assesses the event's likely impact scope, the plausibility of its designated source language, and whether the neighbor remains outside the benchmark-specific boundary. Invalid pairs are revised or returned to the Codex-assisted pairing process, yielding 50 common-goal and 30 language-conditioned pairs.

\subsection{Stage 2: Topic-to-Knowledge-to-QA Transformation}
\label{app:stage2_details}

\paragraph{Knowledge-Unit Schema.}
A knowledge unit represents an atomic, independently queryable fact with a concise answer and directly traceable evidence. Each unit contains a \texttt{relation\_type}, which specifies the general relationship expressed by the fact; a \texttt{semantic\_slot}, which identifies the particular information represented within that relationship; an \texttt{answer} and \texttt{answer\_type}, which define the expected-answer boundary; and a \texttt{fact\_statement} and \texttt{source\_span}, which record the evaluated claim and its supporting evidence. We exclude composite facts, subjective or open-ended claims, facts requiring evidence aggregation across multiple spans, and facts that cannot be answered concisely.

\paragraph{Candidate Extraction and Normalization.}
For each target--neighbor topic pair, Gemini~3.1 Pro Preview first identifies relation types that are represented and directly supported in both topics. 
% Guided by these shared relation types, it then extracts candidate knowledge units separately from the two topics, extracting up to 30 candidate knowledge units from each topic before subsequent filtering. 
Guided by these shared relation types, it then separately extracts up to 30 candidate knowledge units from each topic before filtering.
Each extracted unit must cite a source span from its own topic. We remove units with incomplete fields, unlocatable source spans, duplicate factual statements, or duplicate relation--answer combinations. Units whose answers cannot be directly located in their supporting evidence are flagged for further review.

% \paragraph{Schema-Aware Relation Matching.}
% Gemini~3.1 Pro Preview performs schema-aware matching over the resulting candidate pools. Before matching, temporal and locational expressions are normalized to a common, evidence-supported granularity. A candidate pair must have the same normalized relation type, compatible semantic slots, compatible answer types, and direct evidence in both source topics. Eligible pairs are ranked according to relation and semantic-slot consistency, answer-type compatibility, and evidence quality. Gemini then constructs one-to-one matches without reusing any knowledge unit. In each selected pair, the unit derived from the target topic becomes the forget unit, while the unit derived from the neighboring topic becomes the retain unit.

\paragraph{Schema-Aware Relation Matching.}
Given the candidate pools extracted by Gemini~3.1 Pro Preview, we apply a deterministic schema-aware matcher. Candidate units must have the same normalized relation type, compatible semantic slots and answer types, and direct answer support in their respective source spans. Answer types are compatible when they match exactly or belong to the same predefined family. We rank candidates by answer-type agreement, semantic-slot overlap, source-span support, and extraction priority, and greedily select one-to-one matches without reusing knowledge units or factual statements. Weakly aligned, identical-answer, overlong, or low-scoring candidates are excluded. The target-side unit becomes the forget unit, while the neighbor-side unit becomes the retain unit.

% \paragraph{Coverage-Aware Selection and Review.}
% To ensure balanced topic coverage, we retain exactly ten relation-matched knowledge-unit pairs from each target--neighbor topic pair. Human reviewers compare the eligible candidates and select the ten pairs with the strongest relation consistency, semantic-slot compatibility, answer-type compatibility, and evidence validity. Candidates not selected among these ten pairs are discarded. This procedure yields 500 common-goal and 300 language-conditioned knowledge-unit pairs.

\paragraph{Coverage-Aware Selection and Review.}
To balance topic coverage and pair quality, we apply a coverage-aware selection strategy over the eligible relation-matched candidates. The strategy prioritizes pairs with strong relation consistency, semantic-slot compatibility, answer-type compatibility, and evidence validity while maintaining broad coverage across target--neighbor topic pairs. The resulting selections are manually reviewed, and invalid or ambiguous pairs are removed or replaced with eligible alternatives. This procedure yields 500 common-goal and 300 language-conditioned knowledge-unit pairs.

\paragraph{Holdout-Unit Construction.}
% After fixing the ten forget--retain pairs for each topic pair, we remove all matched units from the corresponding candidate pools and uniformly sample holdout units without replacement from the remaining units. 
After selecting an average of ten forget--retain pairs per topic pair, we remove all matched units from the corresponding candidate pools and uniformly sample holdout units without replacement from the remaining units.
Because the holdout units are drawn from the same topic-specific candidate pools and shared relation inventories as the matched units, they provide closely aligned and challenging non-member examples rather than unrelated random negatives. At the same time, they remain strictly disjoint from the forget and retain sets at the knowledge-unit level. All sampled holdout units undergo the same evidence and validity review as the matched units. Holdout units are used only for membership inference attacks and therefore do not require multiple formulations of the same fact. Each holdout unit is instantiated as one English core QA without surface variants, following the same answer-anchored generation protocol as the matched units.

\paragraph{Core and Surface QA Generation.}
Gemini~3.1 Pro Preview receives only the structured fields of each knowledge unit rather than the full topic content. The forget and retain units in each matched pair are provided jointly so that their QAs remain comparable across the two roles. For each forget or retain unit, Gemini generates one core QA and three semantically equivalent surface variants. Each question is constructed backward from the unit's short answer. Surface variants may alter lexical choices and syntactic structures but must preserve the relation type, semantic slot, expected answer, and supporting evidence.

\paragraph{English-QA Review.}
All 1,600 English QA families generated from the 800 matched knowledge-unit pairs are first reviewed by Codex powered by GPT-5.5 and then individually inspected by one English-proficient human reviewer. Each family contains one core QA and three surface variants. The review examines question validity, answer consistency, preservation of the relation type and semantic slot, evidence grounding, duplicate formulations, potential answer leakage, and cross-role information leakage between the forget and retain QAs. Invalid QAs are corrected or regenerated before entering the multilingual translation stage.

\subsection{Stage 3: Parallel Multilingual Translation}
\label{app:stage3_details}

\paragraph{Dual-Anchor Translation.}
Gemini~3.1 Pro Preview translates each canonical English QA into the other nine benchmark languages. Translation is conditioned on two complementary anchors. The knowledge-unit anchor fixes the underlying fact, relation type, semantic slot, expected-answer boundary, and supporting evidence. The canonical-English-QA anchor fixes the question intent and whether the instance directly asks for the fact as a core QA or accesses the same fact through a surface reformulation. A valid translation must agree with both anchors while changing only its linguistic realization.

\paragraph{Language-Specific Constraints.}
Translations preserve the expected answer semantics while allowing language-appropriate word order and grammatical structure. Names, dates, numbers, abbreviations, and culturally dependent expressions are translated or transliterated according to the conventions of the target language. These adaptations must not broaden or narrow the question scope, change the expected answer type, or collapse the distinction between core and surface formulations.

\paragraph{Blind Back-Translation.}
Gemini~3.1 Pro Preview performs blind back-translation for every translated QA without access to its canonical English counterpart. In a separate verification step, Gemini compares the back-translated QA with the canonical English QA and determines whether they preserve the same question intent and answer semantics. Translations that fail this check are retranslated and verified again.

\paragraph{Independent Cross-Model Review.}
Because Gemini performs both translation and back-translation verification, these stages may share model-specific errors. We therefore use Codex powered by GPT-5.5 as a cross-model verifier. Without access to Gemini's verification decision, Codex reviews every translated QA against the knowledge-unit and canonical-QA anchors. 

\paragraph{Final Human Verification.}
After all model-based reviews and corrections are completed, we conduct a final human audit of 2,000 translated QAs in 20 rounds of 100 instances. We recruit additional multilingual annotators whose language proficiency collectively covers all nine target languages. Each sampled QA is assigned to an annotator familiar with both English and the corresponding target language. The sampled instances cover all target languages, forgetting settings, knowledge roles, and core-versus-surface formulations. Annotators assess whether each translation preserves the underlying fact, question intent, expected-answer boundary, and core-or-surface role while remaining fluent and natural in the target language. The final human audit confirms that all 2,000 sampled QAs satisfy these criteria, consistent with the safeguards built into the preceding pipeline, including schema-aware matching, dual-anchor translation grounded in both the structured knowledge unit and canonical English QA, blind back-translation, and independent cross-model verification. 

% --------------------------------------------------------------------------

% --------------------------------------------------------------------------------
\section{Evaluation Metrics}
\label{app:eval}

\subsection{Reference-Model Aggregation}
In the common-goal setting, Original and Retrain each correspond to a single multilingual model whose parameters do not depend on a source language. After replicating their ten-language evaluation vectors along the source-language dimension, every evaluation language receives the same weight in the Source and Cross macro-averages. Consequently, under our aggregation protocol, the two aggregates are identical for each knowledge role and evaluation metric. In contrast, each row in the language-conditioned setting evaluates a different source-conditioned knowledge set. Its Source and Cross aggregates are therefore not mathematically constrained to be equal.

\subsection{Membership-Inference Protocol}

\paragraph{Attack Setup.}
For each source language $s\in\mathcal{L}$, we treat the forget core QAs used during multilingual knowledge injection as members and the unseen holdout core QAs in the same language as non-members. Let $z=(x,y)$ denote a core QA, where $x$ is the formatted question prompt and $y=(y_1,\ldots,y_T)$ is its tokenized answer. Following the supervised fine-tuning objective, all likelihood-based attack scores are computed over the answer tokens conditioned on the question and preceding answer tokens. We denote the token-level log probability by
\begin{equation}
    \ell_i(z)
    =
    \log \pi_{\theta}
    \left(
        y_i \mid x,y_{<i}
    \right),
\end{equation}
where $\pi_{\theta}$ is the evaluated model. All attack scores are oriented such that a larger score indicates stronger evidence that an example is a member.

\paragraph{Loss Attack.}
The Loss attack exploits the tendency of a model to assign higher likelihood, or equivalently lower negative log-likelihood, to examples encountered during training. We use the negative token-normalized loss as the membership score:
\begin{equation}
    S_{\mathrm{Loss}}(z)
    =
    \frac{1}{T}
    \sum_{i=1}^{T}
    \ell_i(z).
\end{equation}
Thus, examples receiving larger average log probabilities are regarded as more likely to be members.

\paragraph{Min-K\% Attack.}
Min-K\% focuses on the least likely tokens in an example, based on the intuition that unseen examples are more likely to contain low-probability outlier tokens. Let
\begin{equation}
    m
    =
    \max
    \left(
        1,
        \left\lfloor
            \frac{K}{100}T
        \right\rfloor
    \right),
\end{equation}
and let $\mathcal{I}_{K}(z)$ contain the indices of the $m$ answer tokens with the smallest values of $\ell_i(z)$. The Min-K\% score is
\begin{equation}
    S_{\mathrm{Min\text{-}K}}(z)
    =
    \frac{1}{m}
    \sum_{i\in\mathcal{I}_{K}(z)}
    \ell_i(z).
\end{equation}
We use $K=40$ for all models, languages, and unlearning methods.

\paragraph{Min-K\%++ Attack.}
Min-K\%++ calibrates each observed token probability against the complete next-token distribution predicted under the same prefix. Let
\begin{equation}
    p_i(v)
    =
    \pi_{\theta}
    \left(
        v \mid x,y_{<i}
    \right)
\end{equation}
denote the probability assigned to vocabulary token $v\in\mathcal{V}$. The conditional mean and standard deviation of the log probabilities are
\begin{align}
    \mu_i
    &=
    \sum_{v\in\mathcal{V}}
    p_i(v)\log p_i(v), \\
    \sigma_i
    &=
    \sqrt{
        \sum_{v\in\mathcal{V}}
        p_i(v)
        \left(
            \log p_i(v)-\mu_i
        \right)^2
    }.
\end{align}
The standardized score of the observed answer token is
% \begin{equation}
%     r_i(z)
%     =
%     \frac{
%         \ell_i(z)-\mu_i
%     }{
%         \sigma_i+\epsilon
%     },
% \end{equation}
\begin{equation}
r_i(z)
=
\frac{
\ell_i(z)-\mu_i
}{
\sqrt{
\max\left(
\sigma_i^2,
10^{-6}
\right)
}
}.
\end{equation}
% where $\epsilon$ is a small constant for numerical stability. 
Let $\mathcal{I}^{++}_{K}(z)$ contain the indices of the $m$ tokens with the smallest standardized scores. The sequence-level Min-K\%++ score is
\begin{equation}
    S_{\mathrm{Min\text{-}K++}}(z)
    =
    \frac{1}{m}
    \sum_{i\in\mathcal{I}^{++}_{K}(z)}
    r_i(z).
\end{equation}
As with Min-K\%, we set $K=40$ throughout the experiments.

\paragraph{Zlib-Calibrated Attack.}
The Zlib attack calibrates model likelihood using the compression-based complexity of the evaluated answer text. We first define the answer perplexity as
\begin{equation}
    \operatorname{PPL}(z)
    =
    \exp
    \left(
        -
        \frac{1}{T}
        \sum_{i=1}^{T}
        \ell_i(z)
    \right).
\end{equation}
Let $C_{\mathrm{zlib}}(y)$ denote the compressed length, in bytes, of the UTF-8 encoding of the expected-answer text corresponding to $y$ under zlib compression. Following the likelihood-to-compression calibration, we define the member-oriented score as
\begin{equation}
    S_{\mathrm{Zlib}}(z)
    =
    -
    \frac{
        \log \operatorname{PPL}(z)
    }{
        C_{\mathrm{zlib}}(y)
    }.
\end{equation}
The negative sign ensures that larger values correspond to stronger membership evidence: an example is considered more likely to be a member when the model assigns its expected answer unusually high likelihood relative to the answer's compression-based complexity.

\paragraph{ROC-AUC and Cross-Language Aggregation.}
Members are treated as the positive class and holdout examples as the negative class. For attack $a$, we independently compute the raw ROC-AUC in each source language:
\begin{equation}
    \operatorname{AUC}_{a}^{(s)}
    =
    \operatorname{ROC\text{-}AUC}
    \left(
        \left\{
            S_a(z),
            b(z)
        \right\}_{z\in
        \mathcal{Q}^{s,\mathrm{core}}_f
        \cup
        \mathcal{Q}^{s,\mathrm{core}}_h}
    \right),
\end{equation}
where $b(z)=1$ for members and $b(z)=0$ for non-members. We then macro-average over the ten source-language runs:
\begin{equation}
    \widehat{\operatorname{AUC}}_{a}
    =
    \frac{1}{|\mathcal{L}|}
    \sum_{s\in\mathcal{L}}
    \operatorname{AUC}_{a}^{(s)}.
\end{equation}
An AUC of $0.5$ indicates chance-level ordering, whereas a value above $0.5$ indicates that members tend to receive larger attack scores than non-members. A value below $0.5$ indicates reversed ordering, i.e., the forget examples tend to receive lower membership scores than the holdout examples. We retain the raw AUC to preserve this directionality; therefore, values below $0.5$ should not automatically be interpreted as weaker membership leakage.

\subsection{General Multilingual Utility}

\paragraph{Belebele Evaluation.}
\begin{figure*}[htb]
    \centering
    \includegraphics[width=0.9\textwidth]{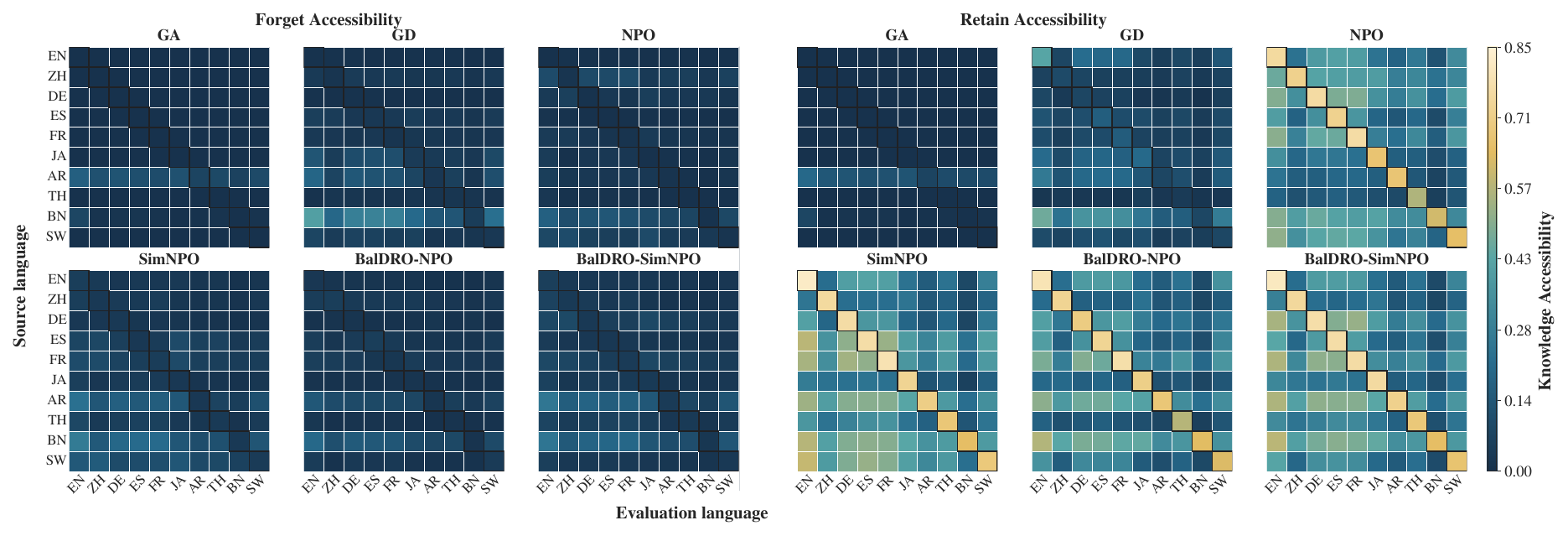}
    \caption{{EM accessibility in the common-goal setting.}
Diagonal cells report source accessibility {\small$A_z^{\mathrm{EM}}(s,s)$}, while off-diagonal cells report cross accessibility {\small$A_z^{\mathrm{EM}}(s,t)$}, where {\small$t\neq s$}. Lower forget accessibility and higher retain accessibility indicate better performance.}
    \label{fig:EM_com}
\end{figure*}

\begin{figure*}[htb]
    \centering
    \includegraphics[width=0.9\textwidth]{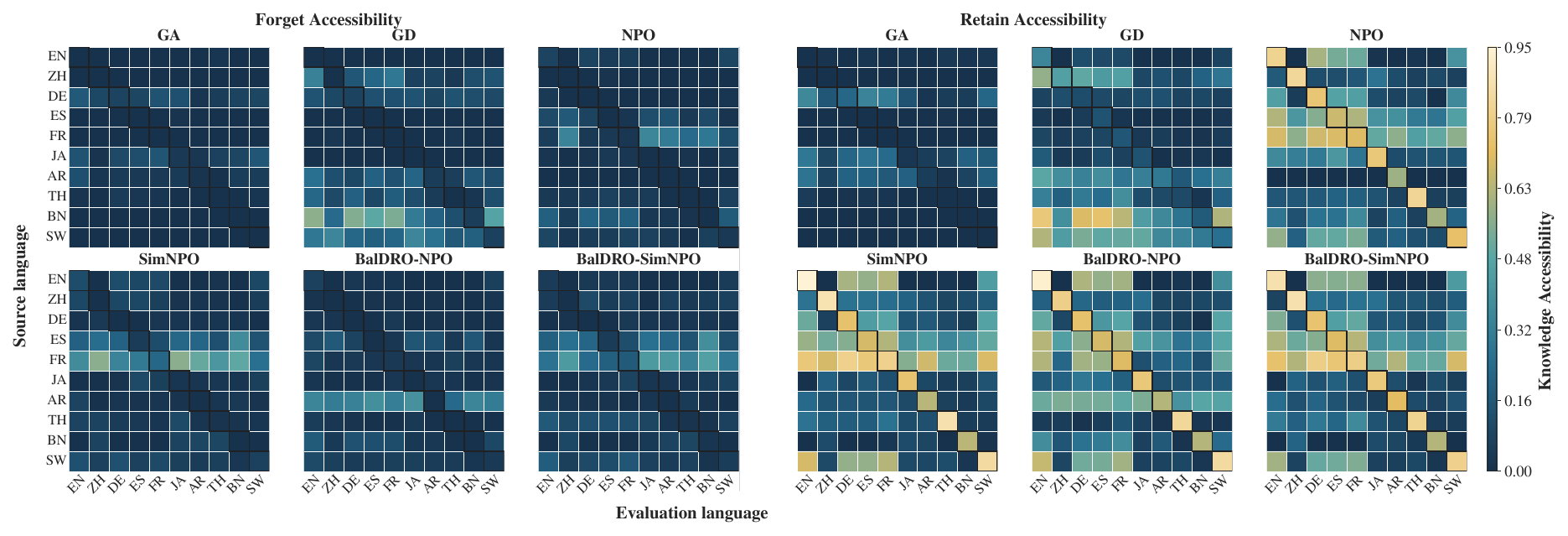}
\caption{{EM accessibility in the language-conditioned setting.}
    Diagonal cells report source accessibility {\small$A_z^{\mathrm{EM}}(s,s)$}, while off-diagonal cells report cross accessibility {\small$A_z^{\mathrm{EM}}(s,t)$}, where {\small$t\neq s$}. Successful unlearning requires low forget accessibility on the diagonal, high forget accessibility off the diagonal, and high retain accessibility across all languages.}
    \label{fig:EM_cul}
\end{figure*}
We evaluate general multilingual utility using Belebele~\cite{bandarkar2024belebele}, a fully parallel multiple-choice reading-comprehension benchmark. For each of the ten benchmark languages, we use all 900 evaluation examples. Each example consists of a passage $c_i$, a question $q_i$, four answer options, and a correct option index $j_i^{*}$.

We evaluate every model in a zero-shot setting using the same model chat template. The passage, question, and four answer options are included in the prompt in the corresponding evaluation language. The model prediction is obtained by selecting the option whose candidate label receives the highest conditional likelihood under the model. Let $\widehat{j}_i$ denote the resulting predicted option index. We report accuracy as the Belebele utility metric.

For an unlearned model obtained using source language $s$ and evaluated in language $t$, its Belebele accuracy is
\begin{equation}
    U(s,t)
    =
    \frac{1}{900}
    \sum_{i=1}^{900}
    \mathbb{I}
    \left[
        \widehat{j}_i^{(s,t)}
        =
        j_i^{*}
    \right].
\end{equation}
We first macro-average over the ten evaluation languages for each source-specific unlearned model:
\begin{equation}
    \widehat{U}(s)
    =
    \frac{1}{|\mathcal{L}|}
    \sum_{t\in\mathcal{L}}
    U(s,t),
\end{equation}
and then macro-average over the ten source-language runs:
\begin{equation}
    \widehat{U}
    =
    \frac{1}{|\mathcal{L}|}
    \sum_{s\in\mathcal{L}}
    \widehat{U}(s).
\end{equation}
For the source-independent Original and Retrain reference models, we evaluate each model once in every language and report the macro-average over the ten evaluation languages. Higher Belebele accuracy indicates better preservation of general multilingual reading-comprehension ability.

\section{Experiments}
\label{app:exp}

\subsection{Experiment Settings}

\paragraph{Details of Evaluated Unlearning Methods.}
We evaluate six representative methods under the same source-language protocol: GA, GD, NPO, SimNPO, BalDRO-NPO, and BalDRO-SimNPO. For each source language {\small$s$}, let {\small$D_f^{(s)}$} and {\small$D_r^{(s)}$} denote the forget and retain sets, respectively. Given an example {\small$z=(x,y)$}, we define the token-normalized retain loss as
\begin{equation}
\ell_r^{(s)}(\theta)
=
\mathbb{E}_{(x,y)\sim D_r^{(s)}}
\left[
-\frac{1}{|y|}
\log\pi_\theta(y\mid x)
\right].
\end{equation}

GA minimizes the reverse cross-entropy on the forget set, while GD additionally minimizes the retain loss~\cite{yao2024large}:
\begin{equation}
\begin{aligned}
\ell_f^{\mathrm{GA},(s)}(\theta)
&=
\mathbb{E}_{(x,y)\sim D_f^{(s)}}
\left[
\frac{1}{|y|}
\log\pi_\theta(y\mid x)
\right],\\
\ell_{\mathrm{all}}^{\mathrm{GA},(s)}(\theta)
&=
\ell_f^{\mathrm{GA},(s)}(\theta),\\
\ell_{\mathrm{all}}^{\mathrm{GD},(s)}(\theta)
&=
\ell_f^{\mathrm{GA},(s)}(\theta)
+
\lambda\ell_r^{(s)}(\theta),
\end{aligned}
\end{equation}
where {\small$\lambda$} controls the retain objective.

\begin{figure*}[htb]
    \centering
    \includegraphics[width=0.9\textwidth]{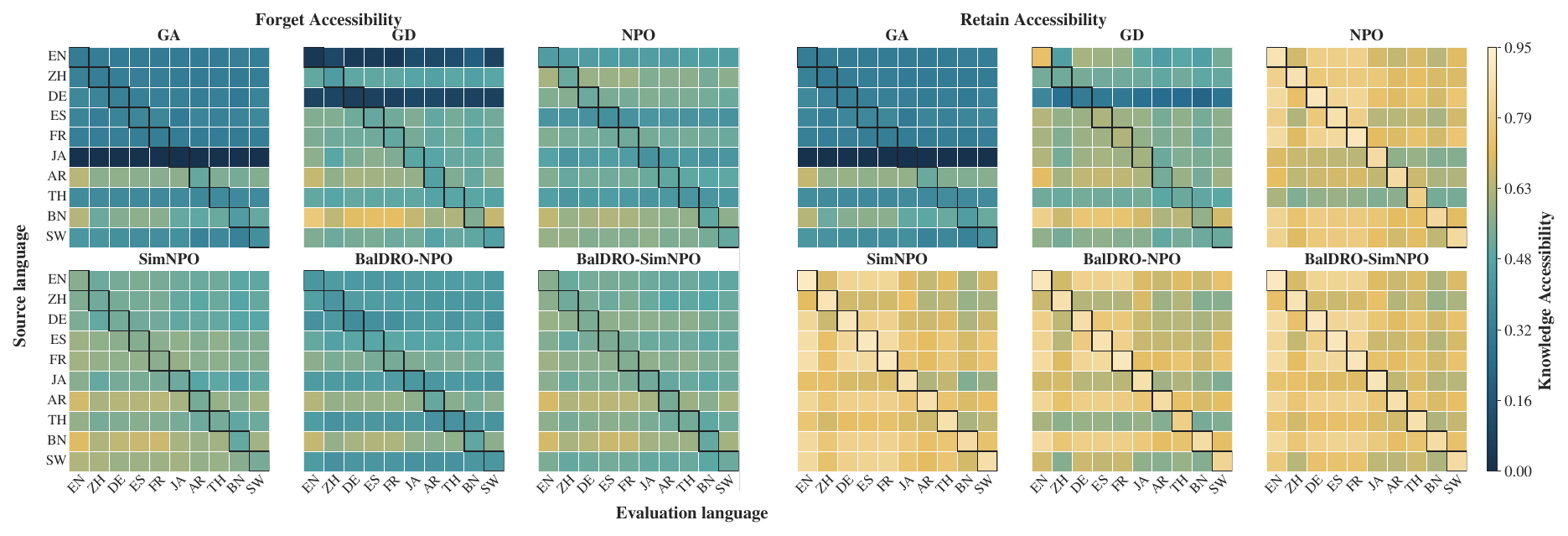}
    \caption{{SS accessibility in the common-goal setting.}
Diagonal cells report source accessibility {\small$A_z^{\mathrm{SS}}(s,s)$}, while off-diagonal cells report cross accessibility {\small$A_z^{\mathrm{SS}}(s,t)$}, where {\small$t\neq s$}. Lower forget accessibility and higher retain accessibility indicate better performance.}
    \label{fig:sim_com}
\end{figure*}

\begin{figure*}[htb]
    \centering
    \includegraphics[width=0.9\textwidth]{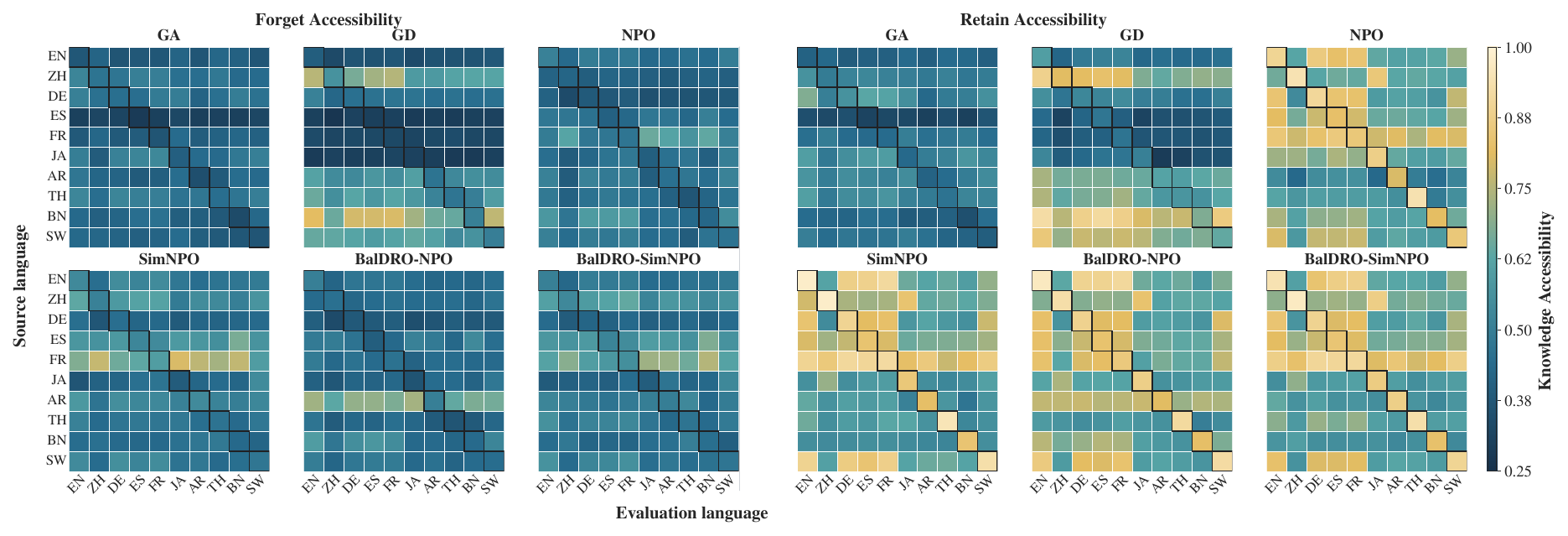}
        \caption{{SS accessibility in the language-conditioned setting.}
    Diagonal cells report source accessibility {\small$A_z^{\mathrm{SS}}(s,s)$}, while off-diagonal cells report cross accessibility {\small$A_z^{\mathrm{SS}}(s,t)$}, where {\small$t\neq s$}. Successful unlearning requires low forget accessibility on the diagonal, high forget accessibility off the diagonal, and high retain accessibility across all languages.}
    \label{fig:sim_cul}
\end{figure*}

NPO uses the frozen Original model {\small$\pi_{\mathrm{ref}}$} to control the decrease in the target-answer probability~\cite{zhang2024negative}. For a forget example {\small$z_i=(x_i,y_i)$}, its per-example loss is
\begin{equation}
\ell_{f,i}^{\mathrm{NPO}}(\theta)
=
-\frac{2}{\alpha^{\mathrm{NPO}}}
\log\sigma\!\left(
-\alpha^{\mathrm{NPO}}
\log
\frac{\pi_\theta(y_i\mid x_i)}
{\pi_{\mathrm{ref}}(y_i\mid x_i)}
\right),
\end{equation}
and the complete objective over a forget batch of size {\small$B$} is
\begin{equation}
\ell_{\mathrm{all}}^{\mathrm{NPO},(s)}(\theta)
=
\frac{1}{B}
\sum_{i=1}^{B}
\ell_{f,i}^{\mathrm{NPO}}(\theta)
+
\lambda\ell_r^{(s)}(\theta).
\end{equation}

% SimNPO removes the reference model and instead uses the length-normalized log-likelihood~\cite{fan2025simplicity}:
% \begin{equation}
% \begin{aligned}
% \ell_{f,i}^{\mathrm{SimNPO}}(\theta)
% &=
% -\frac{2}{\alpha^{\mathrm{Sim}}}
% \log\sigma\!\left(
% -\frac{\alpha^{\mathrm{Sim}}}{|y_i|}
% \log\pi_\theta(y_i\mid x_i)
% \right),\\
% \ell_{\mathrm{all}}^{\mathrm{SimNPO},(s)}(\theta)
% &=
% \frac{1}{B}
% \sum_{i=1}^{B}
% \ell_{f,i}^{\mathrm{SimNPO}}(\theta)
% +
% \lambda\ell_r^{(s)}(\theta).
% \end{aligned}
% \end{equation}
% Here, {\small$\sigma(\cdot)$} denotes the sigmoid function, while {\small$\alpha^{\mathrm{NPO}}$} and {\small$\alpha^{\mathrm{Sim}}$} control the sharpness of the corresponding losses.
SimNPO removes the reference model and applies a margin-based logistic loss to the token-normalized negative log-likelihood~\cite{fan2025simplicity}:
\begin{equation}
\begin{aligned}
\ell_{f,i}^{\mathrm{SimNPO}}(\theta)
&=
-\frac{2}{\alpha^{\mathrm{Sim}}}
\log\sigma\!\left(
\alpha^{\mathrm{Sim}}
\left[
-\frac{1}{|y_i|}
\log\pi_\theta(y_i\mid x_i)
-\delta^{\mathrm{Sim}}
\right]
\right),\\
\ell_{\mathrm{all}}^{\mathrm{SimNPO},(s)}(\theta)
&=
\frac{\gamma^{\mathrm{Sim}}}{B}
\sum_{i=1}^{B}
\ell_{f,i}^{\mathrm{SimNPO}}(\theta)
+
\lambda\ell_r^{(s)}(\theta).
\end{aligned}
\end{equation}
Here, {\small$\sigma(\cdot)$} denotes the sigmoid function, while {\small$\alpha^{\mathrm{NPO}}$} and {\small$\alpha^{\mathrm{Sim}}$}
control the sharpness of the corresponding losses. {\small$\delta^{\mathrm{Sim}}$} specifies the margin applied to the
token-normalized negative log-likelihood, and {\small$\gamma^{\mathrm{Sim}}$} weights the SimNPO forget objective.

BalDRO applies distributionally robust aggregation to the per-example forget losses~\cite{shao2026baldro}. For a batch of losses {\small$\ell_1,\ldots,\ell_B$}, its continuous Donsker--Varadhan aggregator is
\begin{equation}
\mathcal{B}_{\beta}(\ell_1,\ldots,\ell_B)
=
\beta
\log\left(
\frac{1}{B}
\sum_{i=1}^{B}
\exp\left(\frac{\ell_i}{\beta}\right)
\right),
\qquad
\beta>0.
\end{equation}
BalDRO-NPO and BalDRO-SimNPO apply this aggregator to the corresponding per-example forget losses:
\begin{equation}
\begin{aligned}
\ell_{\mathrm{all}}^{\mathrm{BalDRO\text{-}NPO},(s)}(\theta)
&=
\mathcal{B}_{\beta_{\mathrm{NPO}}}
\left(
\ell_{f,1}^{\mathrm{NPO}},\ldots,
\ell_{f,B}^{\mathrm{NPO}}
\right)
+
\lambda\ell_r^{(s)}(\theta),\\
% \ell_{\mathrm{all}}^{\mathrm{BalDRO\text{-}SimNPO},(s)}(\theta)
% &=
% \mathcal{B}_{\beta_{\mathrm{Sim}}}
% \left(
% \ell_{f,1}^{\mathrm{SimNPO}},\ldots,
% \ell_{f,B}^{\mathrm{SimNPO}}
% \right)
% +
% \lambda\ell_r^{(s)}(\theta).
\ell_{\mathrm{all}}^{\mathrm{BalDRO\text{-}SimNPO},(s)}(\theta)
&=
\gamma^{\mathrm{Sim}}
\mathcal{B}_{\beta_{\mathrm{Sim}}}
\left(
\ell_{f,1}^{\mathrm{SimNPO}},\ldots,
\ell_{f,B}^{\mathrm{SimNPO}}
\right)
+
\lambda\ell_r^{(s)}(\theta).
\end{aligned}
\end{equation}
BalDRO is applied only to the forget objective; all preference-based methods use the same token-normalized retain loss. Original and Retrain are reference models rather than unlearning methods.

\paragraph{More Implementation Details.}
We conduct all model-training and evaluation experiments on a server equipped with eight NVIDIA A800 GPUs. Each source-specific unlearning run is assigned to a single GPU. The common-goal runs use a per-device batch size of {\small$8$} with four gradient-accumulation steps, whereas the language-conditioned runs use a per-device batch size of {\small$4$} with two accumulation steps. These configurations produce the effective batch sizes reported in the main text. 

For hyperparameter tuning, in addition to the learning-rate grid reported in the main text, we set {\small$\lambda=1.0$} for every method that uses the retain objective. GA and GD have no additional swept method-specific parameters. For NPO and BalDRO-NPO, we search
{\small$\alpha^{\mathrm{NPO}}\in\{0.1,0.5\}$}. For SimNPO and BalDRO-SimNPO, we search {\small$\alpha^{\mathrm{Sim}}\in\{3.5,4.5\}$} and
{\small$\gamma^{\mathrm{Sim}}\in\{0.125,0.25\}$}, while fixing {\small$\delta^{\mathrm{Sim}}=1.0$}. For the BalDRO variants, we additionally search {\small$\beta_{\mathrm{NPO}}\in\{2,5\}$} and {\small$\beta_{\mathrm{Sim}}\in\{2,5\}$}, respectively. Retain-side robust aggregation is disabled, so both BalDRO parameters apply only to the forget objective. Together with the three learning-rate choices reported in the main text, these settings yield three candidate configurations for GA and GD, six for NPO, twelve for SimNPO and BalDRO-NPO, and twenty-four for BalDRO-SimNPO per source language and setting.

\paragraph{Hyperparameter and Checkpoint Selection.}
Consistent with the evaluation practice of existing LLM unlearning benchmarks, including TOFU~\cite{maini2024tofu} and MUSE~\cite{shi2025muse}, we do not introduce a separate held-out validation split. In many current knowledge unlearning settings, the forget and retain sets directly define the knowledge to be removed and preserved, while the available evaluation examples are constructed around these knowledge units. Consequently, under the fixed knowledge-unit composition and evaluation scope, it is difficult to introduce an additional independent validation split.

During each hyperparameter sweep, we evaluate every epoch checkpoint on the target and neighbor examples in the corresponding source language. Candidate checkpoints are jointly screened using weighted ROUGE-L scores for the target and neighbor examples, with lower target accessibility and higher neighbor accessibility serving as the two selection objectives. Although the hyperparameter search spaces are method-specific, we apply the same checkpoint-selection objectives and screening protocol to all evaluated unlearning methods and source languages. This shared protocol avoids method-specific selection criteria and supports a fair and consistent
comparison among the unlearning methods. The complete hyperparameter ranges and numbers of candidate configurations are reported above.

After checkpoint selection, we use the selected checkpoints to conduct the full multilingual evaluation. The Source results characterize the operating points identified by this common source-language selection protocol, while the complete multilingual results further evaluate the selected checkpoints across different evaluation languages.

\begin{figure}[t]
    \centering
    \includegraphics[width=\columnwidth]{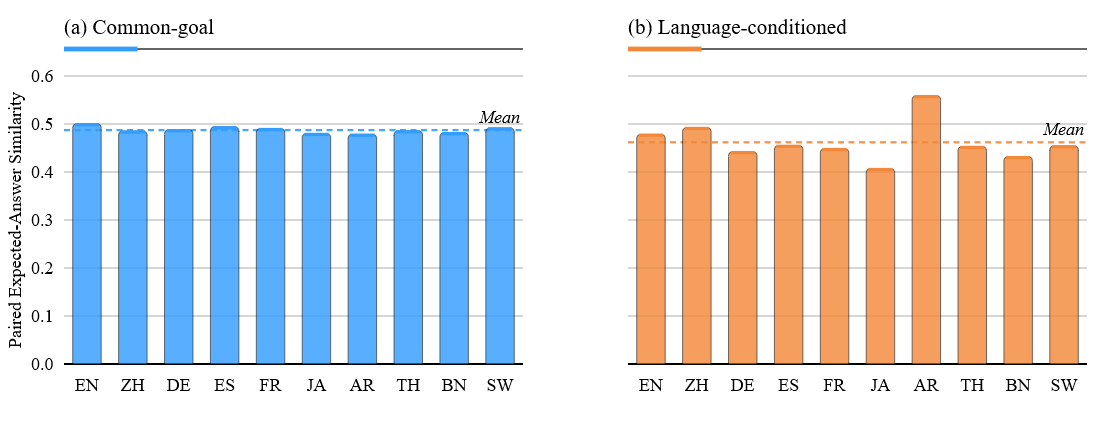}
    \caption{Cosine similarity between the expected answers of paired forget- and retain-set QAs.}
    \label{fig:paired_answer_similarity}
\end{figure}

\paragraph{Metric Analyses.}
SS measures broad semantic proximity rather than factual equivalence. Figure~\ref{fig:paired_answer_similarity} shows that the expected answers of paired forget- and retain-set QAs are semantically similar. Their BGE-M3 similarities reach 0.48--0.50 in the common-goal setting and 0.41--0.56 in the language-conditioned setting. Thus, a non-trivial SS score may reflect proximity to related retain knowledge rather than access to the correct forget knowledge. 

\subsection{More Heatmaps}
\label{app:heat}

To examine fine-grained cross-lingual behavior under different answer-level measures, we visualize knowledge accessibility for every source--evaluation-language pair using EM and SS. Results for the common-goal setting are shown in Figures~\ref{fig:EM_com} and~\ref{fig:sim_com}, while Figures~\ref{fig:EM_cul} and~\ref{fig:sim_cul} present the language-conditioned setting. In each heatmap, diagonal cells represent Source results and off-diagonal cells represent Cross results; the left and right blocks measure forgetting of target knowledge and preservation of neighbor knowledge, respectively.

The results provide different views of forgetting under EM and SS. Many target facts become nearly inaccessible under EM but retain clear semantic signals under SS, showing that failure to produce the exact answer does not imply that the underlying knowledge is fully inaccessible. The methods also show a consistent forgetting utility trade-off: GA and GD generally achieve stronger forgetting but cause greater damage to neighbor knowledge, whereas NPO-based methods preserve more utility but produce weaker and less consistent forgetting across languages. More importantly, the language-conditioned setting rarely exhibits a stable pattern of low diagonal and high off-diagonal forget accessibility. The strong dependence on the source language and transfer direction shows that none of the six evaluated methods can reliably control the cross-lingual boundary of forgetting.

\subsection{LLM-as-a-Judge Accessibility}
\begin{table*}[t]
\centering
\caption{\textbf{LLM-as-a-Judge accessibility under English-source unlearning.}
EN reports evaluation in the source language, while Avg.\ non-EN averages the other nine evaluation languages. These results describe a single English-source run rather than a macro-average over source languages. Arrows indicate the desired direction. }
\label{tab:llm_judge_en}

\resizebox{0.8\textwidth}{!}{
\begin{tabular}{l|cccc|cccc}
\hline

\multirow{3}{*}{\textbf{Method}}
&
\multicolumn{4}{c|}{\textbf{Common-goal}}
&
\multicolumn{4}{c}{\textbf{Language-conditioned}}
\\
\cline{2-5}
\cline{6-9}

&
\multicolumn{2}{c}{\textbf{Forget Accessibility}}
&
\multicolumn{2}{c|}{\textbf{Retain Accessibility}}
&
\multicolumn{2}{c}{\textbf{Forget Accessibility}}
&
\multicolumn{2}{c}{\textbf{Retain Accessibility}}
\\
\cline{2-3}
\cline{4-5}
\cline{6-7}
\cline{8-9}

&
\textbf{EN} $\downarrow$
&
\textbf{Avg.\ non-EN} $\downarrow$
&
\textbf{EN} $\uparrow$
&
\textbf{Avg.\ non-EN} $\uparrow$
&
\textbf{EN} $\downarrow$
&
\textbf{Avg.\ non-EN} $\uparrow$
&
\textbf{EN} $\uparrow$
&
\textbf{Avg.\ non-EN} $\uparrow$
\\
\hline

\textcolor{gray}{Original}
& \textcolor{gray}{0.8238}
& \textcolor{gray}{0.7676}
& \textcolor{gray}{0.8315}
& \textcolor{gray}{0.7747}
& \textcolor{gray}{0.9188}
& \textcolor{gray}{0.8461}
& \textcolor{gray}{0.8896}
& \textcolor{gray}{0.7900}
\\

\textcolor{gray}{Retrain}
& \textcolor{gray}{0.1668}
& \textcolor{gray}{0.1301}
& \textcolor{gray}{0.8261}
& \textcolor{gray}{0.7738}
& \textcolor{gray}{--}
& \textcolor{gray}{--}
& \textcolor{gray}{--}
& \textcolor{gray}{--}
\\
\hline

GA
& 0.0000
& 0.0002
& 0.0015
& 0.0005
& 0.0813
& 0.0875
& 0.1521
& 0.1516
\\

GD
& 0.0114
& 0.0165
& 0.4920
& 0.2611
& 0.0208
& 0.0002
& 0.3542
& 0.2201
\\

NPO
& 0.0379
& 0.0381
& 0.7694
& 0.5958
& 0.1042
& 0.1069
& 0.8083
& 0.7412
\\

SimNPO
& 0.1021
& 0.0907
& 0.8268
& 0.6427
& 0.1417
& 0.1727
& 0.9500
& 0.7725
\\

BalDRO-NPO
& 0.0403
& 0.0517
& 0.8061
& 0.6625
& 0.0896
& 0.0833
& 0.9333
& 0.7889
\\

BalDRO-SimNPO
& 0.0951
& 0.0874
& 0.8231
& 0.6514
& 0.0833
& 0.0789
& 0.8688
& 0.7498
\\
\hline

\end{tabular}
}
\end{table*}
LLM-as-a-Judge requires the evaluator model to score every generated response, making it substantially more expensive than EM, ROUGE-L, and SS. We therefore use only English as the source language and evaluate the resulting English-source Unlearned models in all ten languages. In Table~\ref{tab:llm_judge_en}, EN denotes evaluation in the source language, while Avg.\ non-EN averages the other nine evaluation languages. Since this experiment does not cover all source languages, we report it only as a supplementary diagnostic rather than including it in the ten-source-language macro-average.

The results are consistent with the main evaluation. In the common-goal setting, all methods reduce target accessibility in both English and non-English languages, but GA and GD also cause substantial damage to neighbor knowledge. NPO-based methods preserve more neighbor knowledge at the cost of weaker forgetting. In the {language-conditioned} setting, no method achieves the desired behavior of forgetting in English while preserving the same target knowledge in other languages. This suggests that the main failure is not a general loss of model ability, but the unintended transfer of target forgetting to non-source languages.

\begin{table}[t]
    \centering
    \caption{Topic-level ROUGE-L accessibility in the language-conditioned setting. 
    Values are the mean $\pm$ standard deviation over nine topic pairs. 
    Standard deviations characterize topic-level heterogeneity. 
    % The best mean among the unlearning methods in each column is highlighted in bold.
    }
    \label{tab:topic_level_results}
    \setlength{\tabcolsep}{3pt}
    \renewcommand{\arraystretch}{1.08}
    \resizebox{0.8\columnwidth}{!}{%
        \begin{tabular}{@{}lcccc@{}}
            \toprule
            & \multicolumn{2}{c}{\textbf{Forget Accessibility}}
            & \multicolumn{2}{c}{\textbf{Retain Accessibility}} \\
            \cmidrule(lr){2-3}\cmidrule(l){4-5}
            \textbf{Method}
            & \textbf{Source} $\downarrow$
            & \textbf{Cross} $\uparrow$
            & \textbf{Source} $\uparrow$
            & \textbf{Cross} $\uparrow$ \\
            \midrule
            Original
            & $0.818{\pm}0.134$
            & $0.774{\pm}0.079$
            & $0.829{\pm}0.091$
            & $0.772{\pm}0.097$ \\
            \midrule
            GA
            & $0.037{\pm}0.050$
            & $0.099{\pm}0.074$
            & $0.060{\pm}0.047$
            & $0.141{\pm}0.099$ \\
            GD
            & $\mathbf{0.036}{\pm}0.048$
            & $0.093{\pm}0.131$
            & $0.389{\pm}0.235$
            & $0.197{\pm}0.173$ \\
            NPO
            & $0.056{\pm}0.099$
            & $0.045{\pm}0.039$
            & $0.829{\pm}0.112$
            & $\mathbf{0.286}{\pm}0.096$ \\
            SimNPO
            & $0.098{\pm}0.126$
            & $\mathbf{0.108}{\pm}0.062$
            & $\mathbf{0.882}{\pm}0.105$
            & $0.278{\pm}0.079$ \\
            BalDRO-NPO
            & $0.051{\pm}0.097$
            & $0.045{\pm}0.039$
            & $0.859{\pm}0.122$
            & $0.285{\pm}0.104$ \\
            BalDRO-SimNPO
            & $0.078{\pm}0.086$
            & $0.100{\pm}0.072$
            & $0.861{\pm}0.119$
            & $0.264{\pm}0.088$ \\
            \bottomrule
        \end{tabular}%
    }
\end{table}

\subsection{Per Topic Analyses}

To examine topic-level behavior under the language-conditioned setting, we selected nine target--neighbor topic pairs, with three pairs each from English, Chinese, and Japanese, covering 2,700 core QAs. For each topic, we jointly evaluated the core QAs and their surface variants and aggregated their ROUGE-L accessibility. The intended objective was to suppress the target knowledge in its source language while preserving its accessibility in all other languages; neighbor knowledge should remain accessible across both Source and Cross scopes.

Table~\ref{tab:topic_level_results} shows that none of the evaluated methods established this language boundary. Although Source target accessibility was substantially reduced, Cross target accessibility remained only $0.045$--$0.108$, far below $0.774$ for Original, indicating that forgetting propagated excessively to non-source languages. The preference-based methods preserved source retain accessibility at $0.829$--$0.882$, but their Cross utility decreased to $0.264$--$0.286$, showing that source-language preservation did not generalize across languages. Moreover, the large topic-level deviations, such as $0.093\pm0.131$ for GD on Cross target accessibility and $0.098\pm0.126$ for SimNPO on Source target accessibility, reveal substantial heterogeneity across topics that aggregate results would otherwise conceal.